\documentclass{gensi}
\usepackage{amsmath,amssymb,array,tabularx,xurl}
\usepackage{algorithm,algpseudocode,listings}
\usepackage[percent]{overpic}
\usepackage{tikz}
\usepackage{amsmath,amsfonts,bm}

\def\eqref#1{equation~\ref{#1}}
\def\1{\bm{1}}

\DeclareMathAlphabet{\mathsfit}{\encodingdefault}{\sfdefault}{m}{sl}
\SetMathAlphabet{\mathsfit}{bold}{\encodingdefault}{\sfdefault}{bx}{n}

\newcommand{\method}{EPIC}
\newtheorem{proposition}{Proposition}
\hypersetup{pdftitle={EPIC: Rethinking Trust Regions for Autoregressive LLM Policies via Exact-KL Prefix Importance-Ratio Clipping},pdfauthor={Chengxi Li, Hanlin Wu, Zhicheng Cai, Huan-ang Gao, Wei-Ying Ma, Jingjing Liu, Hao Zhou}}
\title{EPIC: Rethinking Trust Regions for Autoregressive\\
LLM Policies via Exact-KL Prefix Importance-Ratio Clipping}
\author[*,2,3]{Chengxi Li}
\author[*,\ddagger,1,2]{Hanlin Wu}
\author[1,2,3]{Zhicheng Cai}
\author[1,2,3]{Huan-ang Gao}
\author[1,2]{Wei-Ying~Ma}
\author[1,2]{Jingjing Liu}
\author[\dagger,1,2]{Hao Zhou}
\affiliation[1]{SIA-Lab of Tsinghua AIR and ByteDance Seed}
\affiliation[2]{Institute for AI Industry Research (AIR), Tsinghua University}
\affiliation[3]{Department of Computer Science and Technology, Tsinghua University}
\abstract{Reinforcement learning (RL) based on PPO-style clipping has become an effective approach to improving the reasoning capabilities of large language models (LLMs). 
However, standard token-level clipping measures policy change only through the conditional probability of the current token, while the prefix conditioning context is itself generated by the policy.
This mismatch becomes especially consequential in long-response and off-policy training.
We reformulate autoregressive generation as a finite-horizon Markov decision process and derive its performance-difference identity under behavior-policy sampling.
The resulting change of measure identifies the cumulative likelihood ratio of the generated prefix as the exact importance weight for each stepwise policy-improvement term.
Starting from the monotonic-improvement principle of trust-region policy optimization, we then construct a prefix-level trust region using the exact reverse-KL coordinate $r-1-\log r$, whose behavior-policy expectation equals the corresponding prefix-distribution KL divergence.
This yields \method{}, a PPO-style sequence-clipping method that directly regulates prefix-level policy drift without relying on a local quadratic approximation or expert-routing replay.
Experiments on off-policy reinforcement learning with a sparse mixture-of-experts language model show that \method{} consistently outperforms standard and sequence-level clipping baselines across mathematical-reasoning benchmarks, while remaining competitive with routing-replay-based alternatives.
Code is available at \url{https://github.com/LiChengxi666/EPIC}.}
\newcommand{\traininglegend}{%
    \begin{tikzpicture}[baseline]
        \node[
            anchor=south east,
            fill=white,
            fill opacity=0.92,
            text opacity=1,
            draw=black!25,
            rounded corners=1pt,
            inner xsep=2.5pt,
            inner ysep=2pt
        ]{
        {\scriptsize
        \begin{tabular}{@{}ll@{}}
            \textcolor[HTML]{6E6E6E}{\rule{6pt}{1.0pt}}\, GRPO
            &
            \textcolor[HTML]{E34948}{\rule{6pt}{1.0pt}}\, DAPO
            \\[1pt]
            \textcolor[HTML]{EB6834}{\rule{6pt}{1.0pt}}\, GSPO
            &
            \textcolor[HTML]{1BAF7A}{\rule{6pt}{1.0pt}}\, CTPO
            \\[1pt]
            \textcolor[HTML]{4A3AA7}{\rule{6pt}{1.0pt}}\, RIPO
            &
            \textcolor[HTML]{2A78D6}{\rule{6pt}{1.0pt}}\, \method{}
        \end{tabular}}
        };
    \end{tikzpicture}%
}

\begin{document}
\vspace*{5mm}
\maketitle
\begingroup
\renewcommand{\thefootnote}{}
\footnotetext{\textsuperscript{*}Equal contribution.\quad
  \textsuperscript{\(\ddagger\)}Project Lead.\quad
  \textsuperscript{\(\dagger\)}Corresponding author.}
\endgroup
% Allow the overview to float so the introduction can fill the first page.
\begin{figure*}[t]
    \centering
    \captionsetup[subfigure]{position=top,skip=2pt}

    \begin{subfigure}[t]{0.98\textwidth}
        \centering
        \caption{Ratio granularity.}
        \label{fig:prefix-ratio:granularity}
        \includegraphics[
            width=\linewidth
        ]{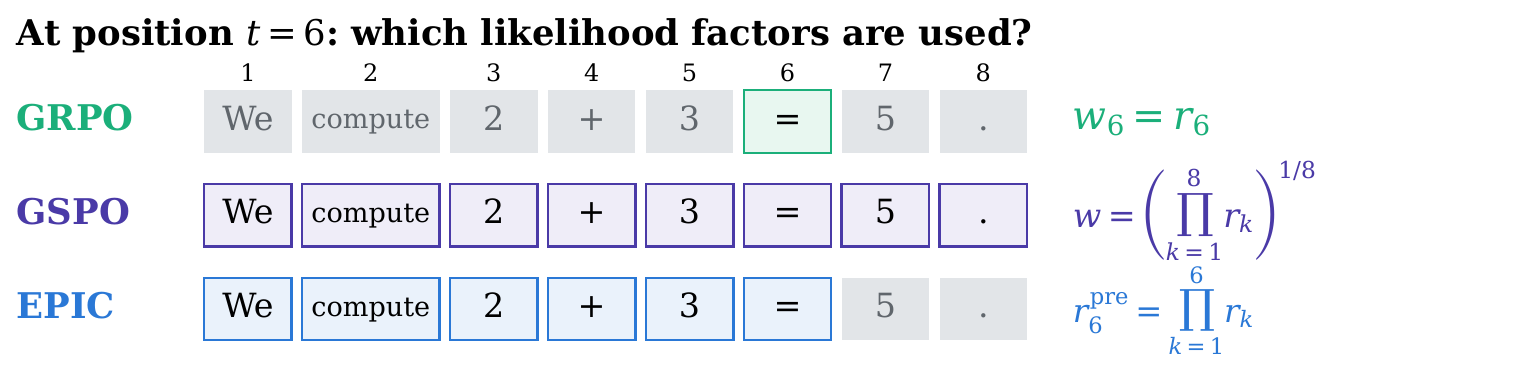}
    \end{subfigure}

    \vspace{3pt}

    \begin{subfigure}[t]{0.50\textwidth}
        \centering
        \caption{Illustrative clipping behavior.}
        \label{fig:prefix-ratio:examples}
        \includegraphics[
            width=\linewidth
        ]{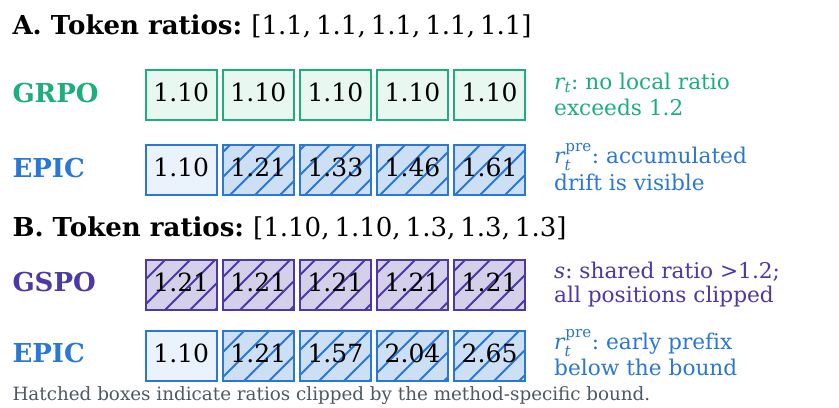}
    \end{subfigure}
    \hfill
    \begin{subfigure}[t]{0.44\textwidth}
        \centering
        \caption{Process memory in use.}
        \label{fig:prefix-ratio:memory}
        \includegraphics[width=\linewidth]
        {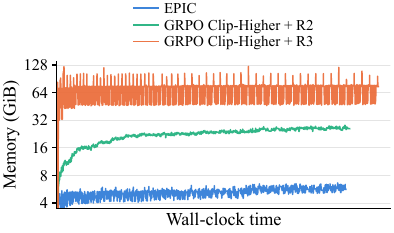}
    \end{subfigure}

    \caption{
        Over view of \method{}.
        \textbf{(a)}
        How GRPO, GSPO, and \method{} compute importance ratios
        along an autoregressive response.
        \textbf{(b)}
        Positive-advantage updates with illustrative clipping bounds.
        GRPO clips token ratios independently, whereas GSPO uses a
        shared response-level ratio and can clip all positions together.
        In these examples, \method{} detects cumulative drift missed
        by GRPO while preserving early-prefix signals suppressed by GSPO.
        Hatched boxes indicate clipped positions.
        \textbf{(c)}
        Driver-process resident CPU memory during training.
        \method{} uses substantially less memory than GRPO Clip-Higher
        combined with R2 or R3, without requiring expert-routing replay.
        The vertical axis is logarithmic.
    }
    \label{fig:prefix-ratio-overview}
\end{figure*}

\section{Introduction}

Reinforcement learning with verifiable rewards (RLVR) has become a central component of
large language model (LLM) post-training, driving substantial advances in reasoning and coding capabilities~\citep{deepseekai2025deepseekr1,yu2025dapo,zheng2025gspo,kimiteam2025kimik15scalingreinforcement}. In throughput-oriented modern LLM RL pipelines such as slime~\citep{slimeteam2025slime} and verl~\citep{sheng2024hybridflow}, fully asynchronous rollout generation and policy optimization~\citep{fu2025areal}, together with differences in kernel implementations between inference and training engines~\citep{zheng2025stabilizingrl}, inevitably introduce distributional mismatch that precludes strictly on-policy training. PPO-style clipping~\citep{schulman2017ppo} is widely adopted to stabilize RL training by discarding gradient contributions associated with large deviations in current-to-behavior probability ratios, which serve as a proxy for policy divergence.

However, even among recently released large-scale models, the appropriate clipping mechanism and granularity for autoregressive LLMs remain unsettled: For example, Ling-2.6~\citep{li2026lingring26}, Qwen-AgentWorld~\citep{zuo2026qwenagentworld}, and Harvey Tenet~\citep{qi2026harveytenet} adopt sequence-level clipping through GSPO,
whereas DeepSeek-V4~\citep{deepseekai2026deepseekv4} and GLM-5~\citep{glm5team2026glm5} continue to use token-level clipping through GRPO.

This unresolved problem is particularly consequential in modern high-throughput LLM training pipelines, where decoupling rollout generation from optimization improves efficiency but introduces policy lag, causing training to rely on samples generated by outdated behavior policies ~\citep{espeholt2018impala,noukhovitch2025asynchronous,li2026staleflow, deepseekai2025deepseekv32}. Existing clipping-based methods remain susceptible to training instability in these off-policy settings ~\citep{zheng2025gspo,zheng2025stabilizingrl}.
To mitigate this instability, current mixture-of-experts (MoE) training pipelines often rely on auxiliary routing-alignment mechanisms, such as Vanilla Routing Replay (R2)~\citep{zheng2025stabilizingrl} and Rollout Routing Replay (R3)~\citep{ma2025r3}, to explicitly suppress the distribution mismatch. However, R2 and R3 limit the freedom of learning and incur additional memory overhead to store expert routing indices, while leaving the more fundamental challenge unresolved: how to stably train LLMs in substantially off-policy settings?

To address this challenge, we revisit the original design and motivation of PPO-clip and reformulate this technique in the autoregressive LLM RL setting. 
TRPO seeks conservative policy improvement by constraining the KL divergence from the behavior policy, while PPO-clip replaces this update with a first-order clipped surrogate~\citep{schulman2015trpo,schulman2017ppo}. 
We formulate autoregressive generation as a finite-horizon MDP and derive its performance-difference identity under behavior-policy sampling.
The resulting change of measure identifies the cumulative prefix ratio as the exact importance weight for each stepwise improvement term~\citep{precup2000eligibility}.
Applying the monotonic-improvement principle of TRPO then leads to a trust region over the behavior- and updated-policy-induced prefix distributions.
Thus, the prefix ratio both measures each autoregressive improvement term and provides the ratio coordinate for its KL constraint.

Motivated by this distinction, we propose \method{}, a prefix-aware clipped objective for off-policy LLM reinforcement learning. 
Based on the prefix policy ratio $r^{\mathrm{pre}}$, we use the exact reverse-KL coordinate $r^{\mathrm{pre}}-1-\log r^{\mathrm{pre}}$, whose expectation under the behavior policy equals the KL divergence between the behavior- and updated-policy-induced prefix distributions.
Combining this exact KL characterization with the monotonic-improvement framework of TRPO, we translate the prefix-level trust region into a practical PPO-style clipped objective.
This construction stabilizes strongly off-policy MoE training without expert-routing replay, thereby avoiding the additional storage and system complexity required to record and replay expert assignments. Figure~\ref{fig:prefix-ratio-overview} illustrates how \method{} differs from token- and response-level clipping and highlights its memory efficiency relative to routing replay.

In conclusion, our contributions are threefold:
\begin{itemize}
    \item We reformulate trust-region policy optimization for autoregressive generation. 
    By deriving the per\-formance-difference identity under behavior-policy sampling, we show that the cumulative prefix ratio is the exact change-of-measure factor for each stepwise improvement term, and extend the TRPO monotonic-improvement principle to policy-induced prefix distributions.

    \item Based on this formulation, we develop \method{}, a practical PPO-style clipped objective that constructs prefix-ratio bounds from
    the exact reverse-KL coordinate
    $r^{\mathrm{pre}}-1-\log r^{\mathrm{pre}}$ and implements the resulting
    prefix-level trust region through efficient token-level computation.

    \item In strongly off-policy MoE training, \method{} outperforms GSPO by $6.50$ percentage points. 
    Compared with routing-replay baselines, \method{} achieves competitive four-benchmark performance while reducing median driver-side process memory by $77.5\%$ and $93.0\%$ relative to R2 and R3.
\end{itemize}

\section{Preliminaries}
We consider a Markov decision process (MDP) $(\mathcal S,\mathcal A,P,R,\rho_0,T)$.
Here $\mathcal S$ and $\mathcal A$ are the state and action spaces, $P(s'\mid s,a)$ is the transition probability, $R(s,a)$ is the reward, $\rho_0$ is the initial-state distribution, and $T$ is the horizon.
A stochastic policy $\pi(a\mid s)$ induces trajectories through $s_0\sim\rho_0$, $a_t\sim\pi(\cdot\mid s_t)$, and $s_{t+1}\sim P(\cdot\mid s_t,a_t)$. The state-visitation distribution at step $t$ is $d_\pi^t(s)=\Pr(s_t=s\mid\pi)$.

\paragraph{Trust Region Policy Optimization (TRPO).}
TRPO motivates trust-region updates with a monotonic policy-improvement bound, aiming to prevent performance collapse caused by large policy changes~\citep{schulman2015trpo}.
Let $\mu=\pi_{\mathrm{old}}$ and $\pi=\pi_\theta$ denote the old and target policies, respectively, and let $J(\pi)=\mathbb E_\pi[\sum_{t=0}^{\infty}\gamma^t R(s_t,a_t)]$ for $\gamma\in(0,1)$.
The performance-difference identity gives~\citep{kakade2002approximately}
\begin{equation}
J(\pi)-J(\mu)
=\sum_s\rho_\pi(s)\sum_a\pi(a\mid s)A^\mu(s,a),
\label{eq:trpo-performance-difference}
\end{equation}
where $\rho_\pi(s)=\sum_{t=0}^{\infty}\gamma^t d_\pi^t(s)$ is the unnormalized discounted state-visitation frequency and $A^\mu$ is the advantage under $\mu$.
Because $\rho_\pi$ changes with $\pi$, old-policy samples do not directly represent the new state visitation~\citep{touati2020stablepolicyoptimizationoffpolicy}.
Replacing $\rho_\pi$ with $\rho_\mu$ yields the tractable improvement surrogate
\begin{equation}
S_\mu(\pi)=\sum_s\rho_\mu(s)\sum_a\pi(a\mid s)A^\mu(s,a).
\label{eq:trpo-improvement-surrogate}
\end{equation}
The original TRPO surrogate is $L_\mu(\pi)=J(\mu)+S_\mu(\pi)$.
Thus, $S_\mu(\pi)$ matches the true improvement in value and policy-parameter gradient at $\pi=\mu$, but can become inaccurate for larger updates.
TRPO controls this approximation error through a policy-divergence penalty, giving the lower bound
\begin{equation}
J(\pi)-J(\mu)
\geq S_\mu(\pi)-C D_{\mathrm{KL}}^{\max}(\mu,\pi)
\mathrel{=:} B_\mu(\pi),
\label{eq:trpo-improvement-bound}
\end{equation}
where $D_{\mathrm{KL}}^{\max}(\mu,\pi)=\max_s D_{\mathrm{KL}}(\mu(\cdot\mid s)\,\|\,\pi(\cdot\mid s))$ and $C\geq0$ depends on the discount factor and a bound on the advantage magnitude.
Since $B_\mu(\mu)=0$, an update with $B_\mu(\pi)\geq0$ guarantees nonnegative improvement under the exact bound.
This motivates constraining the KL divergence at every state: $D_{\mathrm{KL}}^{\max}(\pi_{\mathrm{old}},\pi_\theta)\leq\delta$.
For tractability, TRPO replaces this maximum-KL constraint with an average-KL constraint:
\begin{equation}
\begin{aligned}
\max_\theta\quad&
\mathbb E_{\substack{
t,\ s_t\sim d_{\pi_{\mathrm{old}}}^t,\
a_t\sim\pi_{\mathrm{old}}(\cdot\mid s_t)
}}
\left[\frac{\pi_\theta(a_t\mid s_t)}{\pi_{\mathrm{old}}(a_t\mid s_t)}A_t\right]
\\
\text{s.t.}\quad&
\overline D_{\mathrm{KL}}(\theta)
=\mathbb E_{t,\,s_t\sim d_{\pi_{\mathrm{old}}}^t}
\left[
D_{\mathrm{KL}}
\left(
\pi_{\mathrm{old}}(\cdot\mid s_t)
\,\|\,\pi_\theta(\cdot\mid s_t)
\right)
\right]
\leq\delta,
\end{aligned}
\label{eq:trpo}
\end{equation}
Here $A_t$ is the estimated advantage under the old policy, and $\delta$ is the size of the trust region. TRPO approximately solves this problem by linearizing the objective and using a quadratic approximation to the average KL, requiring curvature information and a line search~\citep{schulman2015trpo}.

\paragraph{Proximal Policy Optimization (PPO).}
PPO-clip seeks conservative policy updates without solving TRPO's KL-constrained subproblem~\citep{schulman2017ppo}.
Define the importance ratio $r_t(\theta)=\pi_\theta(a_t\mid s_t)/\pi_{\mathrm{old}}(a_t\mid s_t)$. With clipping margin $\epsilon$, PPO defines
\begin{equation}
J_{\mathrm{PPO}}(\theta)
=\mathbb E_{\pi_{\mathrm{old}}}
\left[\min\left(r_t(\theta)A_t,\,
\operatorname{clip}(r_t(\theta),1-\epsilon,1+\epsilon)A_t\right)\right].
\label{eq:ppo}
\end{equation}

Actually, these clipping bounds admit a local connection to the quadratic approximation of the trust region condition in Eq.~(\ref{eq:trpo}). 
Requiring every ratio $r_t(\theta)$ to lie in $[1-\epsilon,1+\epsilon]$ would bound the max KL $\max_s D_{\mathrm{KL}}(\pi_{\mathrm{old}}(\cdot\mid s)\,\|\,\pi_\theta(\cdot\mid s))\leq\delta$ by $\epsilon^2/2$ (full proof in Appendix~\ref{app:local-ratio-clipping}).  

\paragraph{MDP for Auto-Regressive Generation.}
Given a prompt $x$, autoregressive language generation can be represented as a finite-horizon MDP~\citep{kwa2024catastrophicgoodhart,bahdanau2017actorcriticalgorithmsequenceprediction}. 
At position $t$, the state consists of the prompt and the previously generated tokens, $s_t=(x,y_{<t})$, while the action is the next token, $a_t=y_t$. The policy is therefore
\begin{equation*}
\pi_\theta(a_t\mid s_t)=\pi_\theta(y_t\mid x,y_{<t}),
\end{equation*}

and taking action $a_t$ deterministically extends the prefix,
so that $s_{t+1}=(x,y_{\leq t})$.

We consider generation up to a fixed horizon $H$. If a response terminates before $H$, it is padded with $\langle\mathrm{eos}\rangle$ tokens that receive zero additional reward, and the terminal value is set to $V^\mu(s_{H+1})=0$. For a fixed prompt $x$, the expected return under policy $\pi$ is
\[
J_x(\pi)
=
\mathbb E_{y\sim\pi(\cdot\mid x)}
\left[
\sum_{t=1}^{H}R(s_t,a_t)
\right].
\]

For the behavior policy $\mu$, define \(Q^\mu(s_t,a_t)\) as the expected remaining return when the current action is fixed to \(a_t\) and all subsequent actions are sampled from
\(\mu\):
\[
Q^\mu(s_t,a_t)
=
\mathbb E\left[
\sum_{k=t}^{H}R(s_k,a_k)
\;\middle|\;
s_t,a_t;\;
a_k\sim\mu(\cdot\mid s_k)\ \text{for }k>t
\right].
\]
The corresponding value and advantage functions are
\begin{equation*}
V^\mu(s_t) = \mathbb E_{a_t\sim\mu(\cdot\mid s_t)}
\left[Q^\mu(s_t,a_t)\right],
\qquad
A^\mu(s_t,a_t) = Q^\mu(s_t,a_t)-V^\mu(s_t).  
\end{equation*}

In autoregressive generation, taking action $a_t=y_t$ deterministically extends the prefix, fixing the next state to $s_{t+1}=(x,y_{\leq t})$. The Bellman identity therefore reduces to

\begin{equation}
\label{eq:autoregressive-advantage-telescope}
Q^\mu(s_t,a_t)=R(s_t,a_t)+V^\mu(s_{t+1}), \quad
A^\mu(s_t,a_t) = R(s_t,a_t) + V^\mu(s_{t+1}) - V^\mu(s_t).
\end{equation}

\section{Proximal Trust Regions For Autoregressive Models}
In this section, we formulate KL-constrained trust-region
optimization for autoregressive language generation. We first derive
an exact performance-difference identity with cumulative prefix
importance weights, then use prefix KL budgets to obtain ratio intervals
and a PPO-style clipped objective.
\subsection{Performance Difference in Autoregressive Generation}
\label{sec:autoregressive-performance-difference}

\paragraph{From advantages to performance difference.}
Eq.~(\ref{eq:autoregressive-advantage-telescope}) holds for any state--action pair, including those generated by $\pi$.
Along each generated trajectory, summing over positions cancels the intermediate value terms:
\begin{equation*}
\begin{aligned}
\sum_{t=1}^H A^\mu(s_t,a_t)
&=\sum_{t=1}^H R(s_t,a_t)
+\sum_{t=1}^H\left[V^\mu(s_{t+1})-V^\mu(s_t)\right]\\
&=\sum_{t=1}^H R(s_t,a_t)+V^\mu(s_{H+1})-V^\mu(s_1)\\
&=\sum_{t=1}^H R(s_t,a_t)-V^\mu(s_1).
\end{aligned}
\end{equation*}
For a fixed prompt, $s_1=(x,\varnothing)$ is fixed and $V^\mu(s_1)=J_x(\mu)$.
Taking the expectation over trajectories sampled from $\pi$, while keeping the advantage defined under $\mu$, gives the finite-horizon counterpart of Eq.~(\ref{eq:trpo-performance-difference}):
\begin{equation}
\begin{aligned}
\mathbb E_{y\sim\pi(\cdot\mid x)}
\left[\sum_{t=1}^H A^\mu(s_t,a_t)\right]
&=\mathbb E_{y\sim\pi(\cdot\mid x)}
\left[\sum_{t=1}^H R(s_t,a_t)-V^\mu(s_1)\right]\\
&=J_x(\pi)-J_x(\mu).
\end{aligned}
\label{eq:autoregressive-performance-difference-on-policy}
\end{equation}

\paragraph{Prefix importance correction.}
To express this identity using behavior-policy samples, assume $\mu(y_{1:t}\mid x)>0$ whenever $\pi(y_{1:t}\mid x)>0$.
The importance weight for a prefix including the current action is
\begin{equation}
r_t^{\mathrm{pre}}
:=\frac{\pi(y_{1:t}\mid x)}{\mu(y_{1:t}\mid x)}
=\prod_{k=1}^t
\frac{\pi(y_k\mid x,y_{<k})}{\mu(y_k\mid x,y_{<k})}.
\label{eq:performance-difference-prefix-weight}
\end{equation}
For each $t$, $A^\mu(s_t,a_t)$ depends only on $x$ and $y_{1:t}$, so a change of measure on the prefix distribution yields
\begin{equation}
\begin{aligned}
\mathbb E_{y\sim\pi(\cdot\mid x)}[A^\mu(s_t,a_t)]
&=\sum_{y_{1:t}}\pi(y_{1:t}\mid x)A^\mu(s_t,a_t)\\
&=\sum_{y_{1:t}}\mu(y_{1:t}\mid x)r_t^{\mathrm{pre}}A^\mu(s_t,a_t)\\
&=\mathbb E_{y\sim\mu(\cdot\mid x)}
\left[r_t^{\mathrm{pre}}A^\mu(s_t,a_t)\right].
\end{aligned}
\label{eq:autoregressive-prefix-change-of-measure}
\end{equation}
Substituting into Eq.~(\ref{eq:autoregressive-performance-difference-on-policy}) gives the exact autoregressive performance-difference identity:
\begin{equation}
J_x(\pi)-J_x(\mu)
=\mathbb E_{y\sim\mu(\cdot\mid x)}
\left[\sum_{t=1}^H r_t^{\mathrm{pre}}A^\mu(s_t,a_t)\right].
\label{eq:autoregressive-performance-difference}
\end{equation}
The prefix weight corrects both the probability of reaching $s_t$ and that of selecting $a_t$; no suffix ratio is needed because the exact advantage already averages future returns under $\mu$.
For the sampled response \(o_i\), we write \(r_{i,t}^{\mathrm{pre}}(\theta) =\pi_\theta(o_{i,1:t}\mid x)/\mu(o_{i,1:t}\mid x)\) for the realization of \(r_t^{\mathrm{pre}}\) under \(\mu=\pi_{\mathrm{old}}\) and \(\pi=\pi_\theta\).

\subsection{Monotonic Improvement Guarantee for AR Model}
\label{sec:autoregressive-trust-region}
Eq.~(\ref{eq:autoregressive-performance-difference}) expresses the true performance improvement using behavior-policy samples.
Taking its right-hand side as the optimization objective gives
\begin{equation}
S_{\mu,x}^{\mathrm{pre}}(\pi)
:=\mathbb E_{y\sim\mu(\cdot\mid x)}
\left[\sum_{t=1}^H r_t^{\mathrm{pre}}A^\mu(s_t,a_t)\right]
=J_x(\pi)-J_x(\mu).
\label{eq:ar-trust-region-objective}
\end{equation}
This plays the role of the surrogate in Eq.~(\ref{eq:trpo-improvement-surrogate}), while retaining the full prefix correction derived in Section~\ref{sec:autoregressive-performance-difference}.
We first use it to construct a TRPO-style monotonic update; Section~\ref{sec:ppo-prefix-clipping} then obtains the corresponding clipped objective.

\paragraph{TRPO-style monotonic improvement for AR policies.}
Following the lower-bound construction in Eq.~(\ref{eq:trpo-improvement-bound}), subtract a nonnegative KL penalty from the improvement objective.
Using the same prefix ratio as in Eq.~(\ref{eq:autoregressive-performance-difference}), write
\begin{equation}
\begin{aligned}
K_{x,t}(\mu,\pi)
&:=\mathbb E_{y\sim\mu(\cdot\mid x)}[-\log r_t^{\mathrm{pre}}]
=D_{\mathrm{KL}}\!\left(\mu(y_{1:t}\mid x)\,\|\,\pi(y_{1:t}\mid x)\right),\\
\mathcal D_x^{\mathrm{pre}}(\mu,\pi)
&:=\max_{1\leq t\leq H}\frac{K_{x,t}(\mu,\pi)}{t}.
\end{aligned}
\label{eq:joint-decision-kl}
\end{equation}
The KL is taken over all length-$t$ prefixes; $\mathcal D_x^{\mathrm{pre}}$ measures the largest cumulative KL per token across prefix lengths.
For $\lambda\geq0$, define
\begin{equation}
B_{\mu,x}^{\mathrm{pre}}(\pi)
:=S_{\mu,x}^{\mathrm{pre}}(\pi)-\lambda\mathcal D_x^{\mathrm{pre}}(\mu,\pi).
\label{eq:ar-prefix-penalized-objective}
\end{equation}
Because the prefix correction already gives the exact improvement, $\lambda$ controls regularization of the update, rather than compensating for a state-visitation approximation.

\begin{proposition}[Monotonic Improvement Condition for AR Model]
\label{prop:ar-monotonic-improvement}
Fix a prompt $x$ and a finite horizon $H$. Assume bounded rewards, common policy support, exact advantages $A^\mu$, and finite $\mathcal D_x^{\mathrm{pre}}(\mu,\pi)$.
For any $\lambda\geq0$,
\begin{equation}
J_x(\pi)-J_x(\mu)\geq B_{\mu,x}^{\mathrm{pre}}(\pi),
\qquad B_{\mu,x}^{\mathrm{pre}}(\mu)=0.
\label{eq:ar-kl-improvement-bound}
\end{equation}
Consequently, any update satisfying $B_{\mu,x}^{\mathrm{pre}}(\pi^+)\geq B_{\mu,x}^{\mathrm{pre}}(\mu)$ guarantees $J_x(\pi^+)\geq J_x(\mu)$.
\end{proposition}

\paragraph{Proof.}
Eq.~(\ref{eq:autoregressive-performance-difference}) and the nonnegativity of KL give
\begin{equation*}
J_x(\pi)-J_x(\mu)-B_{\mu,x}^{\mathrm{pre}}(\pi)
=\lambda\mathcal D_x^{\mathrm{pre}}(\mu,\pi)\geq0.
\end{equation*}
At $\pi=\mu$, both the exact improvement and the KL penalty vanish, so $B_{\mu,x}^{\mathrm{pre}}(\mu)=0$.
Thus $B_{\mu,x}^{\mathrm{pre}}(\pi^+)\geq0$ implies nonnegative performance improvement.\hfill$\square$

This yields the TRPO-style monotonic-update problem
\begin{equation}
\pi^+\in\arg\max_\pi B_{\mu,x}^{\mathrm{pre}}(\pi),
\qquad
J_x(\pi^+)-J_x(\mu)
\geq B_{\mu,x}^{\mathrm{pre}}(\pi^+)
\geq B_{\mu,x}^{\mathrm{pre}}(\mu)=0,
\label{eq:ar-monotonic-update}
\end{equation}
where the policy class is assumed to contain $\mu$.
As in TRPO, the penalized objective can instead motivate a KL-constrained update with a tunable budget:
\begin{equation}
\max_\pi\quad S_{\mu,x}^{\mathrm{pre}}(\pi)
\qquad\text{s.t.}\quad
\mathcal D_x^{\mathrm{pre}}(\mu,\pi)\leq\delta
\quad\Longleftrightarrow\quad
K_{x,t}(\mu,\pi)\leq t\delta\quad\forall t.
\label{eq:ar-trust-region-update}
\end{equation}
For fixed $\lambda$, its Lagrangian is $B_{\mu,x}^{\mathrm{pre}}(\pi)+\lambda\delta$.
The behavior policy is feasible with objective zero, so an exact constrained maximizer also has nonnegative true improvement.
The remaining question is how to replace this explicit KL-constrained optimization by a clipped objective.

\subsection{PPO-Style Prefix Clipping}
\label{sec:ppo-prefix-clipping}
We translate the KL-constrained update in Eq.~(\ref{eq:ar-trust-region-update}) into a PPO-style objective that can be optimized with the existing policy-gradient training loop.

\paragraph{From the KL constraint to clipping bounds.}
Under common support, $\mathbb E_\mu[r_t^{\mathrm{pre}}]=1$.
Therefore the constraint can be expressed through a nonnegative ratio coordinate:
\begin{equation}
\begin{aligned}
K_{x,t}(\mu,\pi)
&=\mathbb E_\mu[-\log r_t^{\mathrm{pre}}]
+\mathbb E_\mu[r_t^{\mathrm{pre}}-1]\\
&=\mathbb E_\mu[\phi(r_t^{\mathrm{pre}})] \\
&=\sum_{o_{1:t}}\mu(o_{1:t}\mid x)\,
\phi\!\left(r_t^{\mathrm{pre}}\right),
\qquad
\phi(r)=r-1-\log r.
\end{aligned}
\label{eq:ar-kl-coordinate}
\end{equation}
Thus, the contribution of a particular prefix \(o_{1:t}\) to the distribution-level KL is weighted by its behavior-policy probability $\mu(o_{1:t}\mid x)$~\citep{cai2026ripo}.
To retain the per-token expected KL-budget target of $\delta$ established in Eq.~(\ref{eq:ar-trust-region-update}), we use the geometric mean $[\mu(o_{i,1:t}\mid x)]^{1/t}$ as the probability-calibration factor.
Following the budget in Eq.~(\ref{eq:ar-trust-region-update}), we impose the calibrated pointwise restriction $\left[\mu(o_{i,1:t}\mid x)\right]^{1/t}\phi(r_t^{\mathrm{pre}}) \leq t\delta$ for every prefix at position $t$. 
The lower- and upper-side coordinate budgets are therefore
\begin{equation}
\kappa^-_{i,t}=\frac{t\delta^-}{[\mu(o_{i,1:t}\mid x)]^{1/t}},
\qquad
\kappa^+_{i,t}=\frac{t\delta^+}{[\mu(o_{i,1:t}\mid x)]^{1/t}}.
\label{eq:calibrated-prefix-budget}
\end{equation}
The factor \(t\delta^\pm\) retains the linear prefix-length scaling implied by \(K_{x,t}(\mu,\pi)\leq t\delta\) in Eq.~(\ref{eq:ar-trust-region-update}), while the geometric-mean behavior probability provides the sample-dependent calibration. Since $\phi$ decreases on $(0,1]$ and increases on $[1,\infty)$, this defines a ratio interval:
\begin{equation}
\phi(\ell_{i,t}) = \kappa^-_{i,t},
\quad
\phi(u_{i,t}) = \kappa^+_{i,t},
\qquad
0<\ell_{i,t}\leq1\leq u_{i,t}.
\label{eq:calibrated-prefix-endpoints}
\end{equation}

\paragraph{The clipped improvement objective.}
We now use these bounds to remove the incentive for updates that improve an individual advantage term by moving its ratio farther outside the interval, following PPO~\citep{schulman2017ppo}.
The prefix ratio can be computed from the token log probabilities already evaluated in a PPO-style training loop. For a sampled response \(o_i=(o_{i,1},\ldots,o_{i,T_i})\),
\begin{equation}
\log r^{\mathrm{pre}}_{i,t}(\theta)
=
\sum_{k=1}^{t}
\left[
\log\pi_\theta(o_{i,k}\mid x,o_{i,<k})
-
\log\mu(o_{i,k}\mid x,o_{i,<k})
\right].
\end{equation}
A cumulative sum obtains the prefix log-ratios at all positions in \(O(T_i)\) work. This calculation requires no additional model pass, rollout, auxiliary network, or vocabulary-wide KL evaluation. These properties make prefix-level clipping compatible with the existing rollout, reward, batching, and optimization pipeline. 

For token-level optimization, we use the geometric mean $g_{i,t}(\theta)=[r^{\mathrm{pre}}_{i,t}(\theta)]^{1/t}$ to place the cumulative prefix ratio on a length-normalized scale and mitigate length-dependent gradient spikes, as supported by the ablation in Appendix~\ref{app:geomprob-ablation}. 
We apply the same transformation to the lower and upper bounds, yielding $\ell_{i,t}^{1/t}$ and $u_{i,t}^{1/t}$, respectively.
To prevent an extreme ratio from producing an unbounded negative contribution, we use the sign-aware dual-clipping operator~\citep{ye2020masteringcomplexcontrolmoba}
\begin{equation}
\mathcal C(q,A;\ell,u)
=
\begin{cases}
A\min(q,u), & A\geq 0,\\
A\min\!\bigl(\max(q,\ell),u\bigr), & A<0.
\end{cases}
\label{eq:dual-clipping-operator}
\end{equation}
Using the straight-through weight $\widetilde g_{i,t}(\theta)$, whose forward value is $g_{i,t}(\theta)$ while its gradient passes only through the current-token probability, the final training objective is
\begin{equation}
\mathcal J_{\mathrm{\method{}}}(\theta)
=
\mathbb E_{\substack{
x\sim\mathcal D\\
\{o_i\}_{i=1}^{G}\sim\mu(\cdot\mid x)
}}
\left[
\frac{1}{G}
\sum_{i=1}^{G}
\frac{1}{T_i}
\sum_{t=1}^{T_i}
\mathcal C\!\left(
\widetilde g_{i,t}(\theta),
\widehat A_i;
\ell_{i,t}^{1/t},
u_{i,t}^{1/t}
\right)
\right].
\label{eq:exact-prefix-objective}
\end{equation}

\begin{algorithm}[t]
\caption{\method{} for Off-Policy Autoregressive RL}
\label{alg:essencerl}
\begin{algorithmic}[1]
\Require Policy $\pi_\theta$, training distribution $\mathcal D$,
budgets $\delta^-,\delta^+$, group size $G$, update depth $K$
\For{each training iteration}
    \State Sample prompts $x\sim\mathcal D$ and set $\mu\gets\pi_\theta$
    \State Generate response groups
    $\{o_i\}_{i=1}^{G}\sim\mu(\cdot\mid x)$
    \State Compute verifier rewards, group-relative advantages
    $\widehat A_i$, and behavior log-probabilities
    \For{$k=1,\ldots,K$}
        \State Evaluate current-policy token log-probabilities
        \State Compute prefix ratios
        $r^{\mathrm{pre}}_{i,t}
        =\prod_{j=1}^{t}
        \pi_\theta(o_{i,j}\mid x,o_{i,<j})/
        \mu(o_{i,j}\mid x,o_{i,<j})$
        \State Set
        $g_{i,t}=(r^{\mathrm{pre}}_{i,t})^{1/t}$
        and
        $\bar p^{\mathrm{old}}_{i,t}
        =[\mu(o_{i,1:t}\mid x)]^{1/t}$
        \State Solve
        $\phi(\ell_{i,t})=t\delta^-/\bar p^{\mathrm{old}}_{i,t}$
        and
        $\phi(u_{i,t})=t\delta^+/\bar p^{\mathrm{old}}_{i,t}$
        for $\phi(r)=r-1-\log r$
        \State Transform bounds to
        $\ell_{i,t}^{1/t}$ and $u_{i,t}^{1/t}$
        \State Form the straight-through weight $\widetilde g_{i,t}$
        \State Compute
        $\mathcal S_{i,t}
        =
        \mathcal C(\widetilde g_{i,t},\widehat A_i;
        \ell_{i,t}^{1/t},u_{i,t}^{1/t})$
        \State Update $\theta$ using
        $\frac{1}{G}\sum_i\frac{1}{T_i}\sum_t\mathcal S_{i,t}$
    \EndFor
\EndFor
\end{algorithmic}
\end{algorithm}

\section{Experiments}
\subsection{Experimental Setup}

\textbf{Models and datasets.}
We train Qwen3-30B-A3B-Base~\citep{yang2025qwen3technicalreport} \footnote{\url{https://huggingface.co/Qwen/Qwen3-30B-A3B-Base}}, a sparse MoE language model, on the processed DAPO-Math-17K dataset~\citep{yu2025dapo}.
A rule-based verifier provides binary answer-correctness scores.
We evaluate on four held-out mathematical-reasoning benchmarks: AIME 2024, AIME 2025, AMC 2023, and HMMT.

\textbf{Baselines.}
We compare \method{} with GRPO, DAPO, GSPO, CTPO, and
RIPO~\citep{shao2024deepseekmathpushinglimitsmathematical,
yu2025dapo,zheng2025gspo,zhang2026ctpo,cai2026ripo}.
The comparisons use the same training and evaluation protocol.
Baselines use their published default clipping parameters, while \method{} uses $\delta^-=5\times10^{-4}$ and $\delta^+=2\times10^{-3}$.
We additionally compare against GRPO Clip-Higher combined with
Vanilla Routing Replay (R2) or Rollout Routing Replay (R3)%
~\citep{zheng2025gspo,zheng2025stabilizingrl,ma2025r3}.

\textbf{Training and evaluation.}
Each run is trained under the same protocol.
We evaluate before training and every five global steps, sampling eight responses per problem and reporting $\mathrm{mean@8}$, their average binary correctness.
For each run, we select the single checkpoint that maximizes the equal-weight average across the four benchmarks and report all four scores at that checkpoint. 
Main results are averaged over three independent runs.
Dataset details and training configurations are provided in Appendix~\ref{app:experimental-config}.

\subsection{Main Results}

\begin{figure*}[t]
    \centering
    \captionsetup[subfigure]{position=top,skip=2pt}

    \begin{subfigure}[t]{0.43\textwidth}
        \centering
        \caption{Validation performance.}
        \label{fig:dynamics:validation}
        \begin{overpic}[width=\linewidth]
            {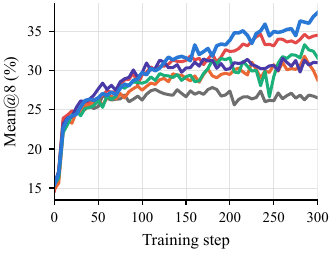}
            \put(97,20){%
                \makebox(0,0)[br]{\traininglegend}%
            }
        \end{overpic}
    \end{subfigure}
    \hfill
    \begin{subfigure}[t]{0.43\textwidth}
        \centering
        \caption{Training reward.}
        \label{fig:dynamics:reward}
        \begin{overpic}[width=\linewidth]
            {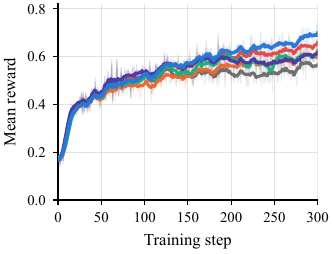}
            \put(97,20){%
                \makebox(0,0)[br]{\traininglegend}%
            }
        \end{overpic}
    \end{subfigure}

    \caption{
        Validation and reward trajectories under the aligned off-policy
        MoE setting. Curves average three seeds at matching training steps.
    }
    \label{fig:training-dynamics}
\end{figure*}

Table~\ref{tab:main-results} reports the main comparison under an
update depth of eight. For each method, we select the single checkpoint
that maximizes the equal-weight arithmetic mean of the four validation
\(\mathrm{mean@8}\) scores. 
\method{} achieves the highest four-benchmark average of $39.34\%$, exceeding GRPO and GSPO by $10.69$ and $6.50$ percentage points, respectively.
The improvements are particularly pronounced on AIME 2024 and HMMT 2025, while the gain on AMC 2023 indicates that performance on the more difficult benchmarks is not obtained at the expense of the relatively easier one.
Relative to GSPO, the gains are $8.06$, $4.72$, $4.48$, and $8.75$ percentage points on AIME 2024, AIME 2025, AMC 2023, and HMMT 2025, respectively. This consistent direction across benchmarks shows that the aggregate improvement is not driven by a single evaluation set.
Figure~\ref{fig:training-dynamics}\subref{fig:dynamics:validation} shows the four-benchmark average throughout training.
The advantage becomes clearer during later training, complementing the selected-checkpoint comparison with evidence from the learning trajectories.
Figure~\ref{fig:training-dynamics}\subref{fig:dynamics:reward} further shows that \method{} achieves the highest late-stage training reward among the compared methods.
Individual benchmark curves and the corresponding analysis are provided in Appendix~\ref{app:validation-trajectories}.

\begin{table}[t]
\caption{Main results on Qwen3-30B-A3B-Base. Results are averaged over three independent runs.}
\label{tab:main-results}
\centering
\small
\setlength{\tabcolsep}{5.2pt}
\renewcommand{\arraystretch}{1.12}
\begin{tabular}{lrrrrr}
\toprule
& \multicolumn{5}{c}{Accuracy (\%)} \\
\cmidrule(lr){2-6}
Method
& AIME 24
& AIME 25
& AMC 23
& HMMT 25
& Average \\
\midrule

GRPO
& 23.61
& 16.81
& 69.48
& 4.72
& 28.65
\\

DAPO
& 31.67
& 24.03
& 76.25
& 9.86
& 35.45
\\

GSPO
& 30.00
& 20.56
& 73.44
& 7.36
& 32.84
\\

CTPO
& 30.14
& 20.97
& 76.04
& 8.75
& 33.98
\\

RIPO
& 26.94
& 21.81
& 74.69
& 8.47
& 32.98
\\

\midrule

\textbf{\method{} (Ours)}
& \textbf{38.06}
& \textbf{25.28}
& \textbf{77.92}
& \textbf{16.11}
& \textbf{39.34}
\\

\bottomrule
\end{tabular}
\end{table}

\subsection{Trust region design analysis}

\begin{figure*}[t]
    \centering
    \captionsetup[subfigure]{position=top,skip=2pt}

    \begin{subfigure}[t]{0.32\textwidth}
        \centering
        \caption{Position profile.}
        \label{fig:trust-region:profile}
        \includegraphics[width=\linewidth]
        {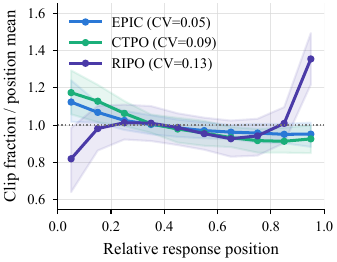}
    \end{subfigure}
    \hfill
    \begin{subfigure}[t]{0.32\textwidth}
        \centering
        \caption{Position-wise CV.}
        \label{fig:trust-region:cv}
        \includegraphics[width=\linewidth]
        {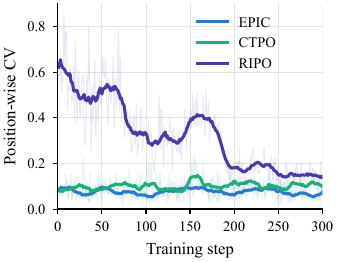}
    \end{subfigure}
    \hfill
    \begin{subfigure}[t]{0.32\textwidth}
        \centering
        \caption{Probability calibration.}
        \label{fig:trust-region:calibration}
        \includegraphics[width=\linewidth]
        {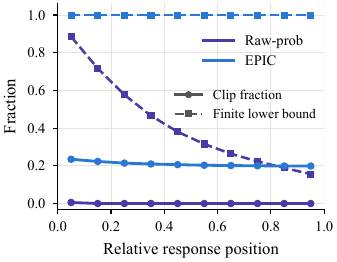}
    \end{subfigure}

    \caption{
            Position-wise behavior of the calibrated trust region. \textbf{(a)} Normalized late-stage clip fractions across relative response positions. \textbf{(b)} Position-wise clip-fraction variation during training. \textbf{(c)} Clip fraction and finite-lower-bound coverage under raw and geometric-mean behavior-probability calibration.
    }
    \label{fig:trust-region-analysis}
\end{figure*}

Here we examine whether the proposed trust-region design produces effective and balanced clipping throughout autoregressive generation.
Figure~\ref{fig:trust-region-analysis}\subref{fig:trust-region:profile} shows that \method{} achieves a late-stage position-wise coefficient of variation (CV) of $0.054$, compared with $0.085$ for CTPO and $0.126$ for RIPO.
Figure~\ref{fig:trust-region-analysis}\subref{fig:trust-region:cv}
further shows that its position-wise CV remains consistently low
throughout training.
Together, these observations indicate that clipping activity remains
relatively evenly distributed across response positions rather than
becoming increasingly concentrated as training progresses.
Figure~\ref{fig:trust-region-analysis}\subref{fig:trust-region:calibration}
reports the Raw-prob versus default comparison from the ablation
study in Appendix~\ref{app:geomprob-ablation}.
Using raw prefix probabilities concentrates clipping near the
beginning of the response, whereas our calibration maintains finite
lower bounds across all position bins and clip fractions of
approximately $0.20$--$0.23$.
This contrast illustrates the role of calibration in preventing
later prefixes from becoming effectively unclipped.
The normalized position profiles and absolute clip fractions provide complementary diagnostics: the former measures how clipping is distributed within a response, while the latter checks whether clipping remains active at all. A low position-wise CV alone would not establish the latter.
Overall, these results support the proposed design by showing
that clipping remains active across response positions rather
than concentrating on a small portion of the sequence.

\subsection{Comparison with Routing Replay}
\label{sec:routing-replay}

% Keep the comparison figure beside its discussion, before the conclusion page break.
\begin{figure}[!htbp]
    \centering
    \captionsetup[subfigure]{position=top,skip=2pt}

    \begin{subfigure}[t]{0.55\linewidth}
        \centering
        \caption{Benchmark performance.}
        \label{fig:routing-replay:benchmark}
        \includegraphics[width=\linewidth]
        {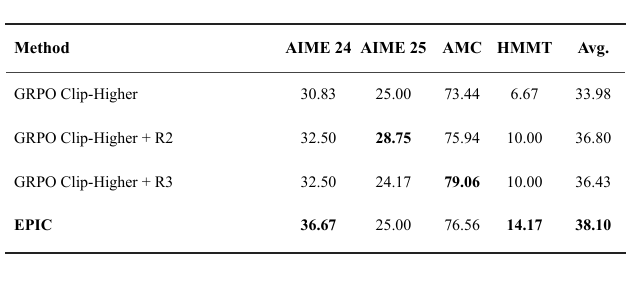}
    \end{subfigure}
    \hfill
    \begin{subfigure}[t]{0.41\linewidth}
        \centering
        \caption{Process memory in use.}
        \label{fig:routing-replay:memory}
        \includegraphics[width=\linewidth]
        {figures/router_replay_process_memory_replay_only.pdf}
    \end{subfigure}

    \caption{
        Comparison with routing replay.
        \textbf{(a)} Mean@8 accuracy (\%) at the checkpoint maximizing
        the equal-weight average across the four benchmarks.
        \textbf{(b)} Process-level resident CPU memory during training.
    }
    \label{fig:routing-replay-comparison}
\end{figure}

We compare \method{} with GRPO with higher upper clip combined with R2 and R3 under the same protocol.
Unlike these replay-based approaches, \method{} does not store or replay expert-routing decisions and modifies only the clipped policy objective.
As shown in Figure~\ref{fig:routing-replay:benchmark}, both R2 and R3 substantially improve GRPO Clip-Higher in this off-policy setting.
\method{} slightly outperforms both replay-based variants on the four-benchmark average without replaying expert-routing decisions.
Figure~\ref{fig:routing-replay:memory} shows that \method{} maintains a substantially lower and more stable process-memory usage than GRPO Clip-Higher with R2 or R3 throughout the recorded training period.
As these methods have similar response lengths (\(8{,}557\), \(9{,}476\), and \(9{,}001\) tokens for R2, R3, and \method{}), these measurements suggest that \method{} avoids substantial driver-side memory overhead associated with carrying routing-replay state.
Together with its competitive benchmark performance, these observations suggest that \method{} offers a memory-efficient alternative to routing replay in the evaluated setting.
The reported memory metric is driver-process resident CPU memory, rather than GPU memory or total cluster memory. It characterizes the observed system-level cost, not an isolated measurement of routing-state storage.

\section{Conclusion}
We revisit PPO-style clipping through a finite-horizon MDP formulation of autoregressive generation, identifying the prefix ratio as the exact stepwise importance weight.
Building on TRPO's monotonic-improvement framework, we derive clipping bounds from the exact reverse-KL coordinate to obtain \method{}.
Experiments show improved reasoning performance over clipping baselines and competitive results with routing replay at lower process-memory cost.
This work connects trust-region theory with practical optimization, advancing principled and efficient reinforcement learning for language models.

\clearpage
\subsection*{AI use statement}

Generative AI tools were used to assist with theoretical derivations,
language editing, clarification of exposition, literature-search
formulation, code implementation, environment configuration, experiment
execution, exploratory data analysis, visualization scripts, and LaTeX
formatting. AI tools were not used to generate experimental results,
fabricate or alter data, make final scientific decisions, or replace
author verification of the theoretical derivations and empirical claims.

All AI-assisted text, derivations, code, configurations, experiment procedures, figures, and analytical suggestions were reviewed and revised by the authors. 
The authors independently checked the
mathematical derivations, verified the numerical results against the recorded experiment logs, inspected the generated figures, and tested the analysis code used to produce the reported plots. The final paper, including all claims, equations, experimental results, and artifacts, is the responsibility of the authors.
We take responsibility for the final content of this work, including text, claims or artifacts produced with the aid of generative AI.

\subsection*{Ethics statement}

This work studies optimization methods for reinforcement learning
post-training of large language models. It does not involve human
subjects, human-subject data, personally identifiable information,
or the collection of sensitive personal data. The experiments use
language-model training and mathematical-reasoning benchmark
evaluation in an automated setting.

We are not aware of any immediate ethical risks specific to the
methods or experiments presented here. As with other research on
large language models, the resulting models may inherit limitations
or biases from their training data and should not be deployed in
high-stakes settings without appropriate evaluation, monitoring, and
safeguards.

\subsection*{Reproducibility Statement}

Our results are based on publicly available language models, datasets,
and experimental frameworks. The experiments use publicly released
model checkpoints and benchmark datasets, together with the publicly
available training framework described in the experimental setup. The
experimental section reports the model setting, benchmark suite,
training configuration, evaluation protocol, and multi-seed results.

We provide the theoretical assumptions, derivations, and implementation
details needed to reproduce the proposed objective in the main paper
and appendix. The appendix includes the complete prefix-ratio
construction, the reverse-KL identity, the trust-region budget
calibration, the clipping objective, and the additional implementation
details used in training. In particular, the experimental configurations
and hyperparameters are documented in Appendix~\ref{app:experimental-config}
while the additional analyses and diagnostic procedures are described in
Appendix~\ref{app:further-analysis}.

The method implementation, training configurations, and instructions for environment setup, training, and evaluation are publicly available at \url{https://github.com/LiChengxi666/EPIC}.
The repository builds on verl and releases the EPIC additions under the Apache-2.0 license, with upstream notices retained.

\clearpage
\bibliographystyle{unsrtnat}
\bibliography{references}

\clearpage
\beginappendix

\section{Related work}
\paragraph{Importance Sampling in Off-Policy Reinforcement Learning.}
In off-policy reinforcement learning, importance sampling provides a standard mechanism for estimating expectations under a target policy using samples collected by a different behavior policy~\citep{degris2013offpolicyactorcritic}. 
In sequential decision problems, 
a full-trajectory likelihood ratio gives an unbiased correction 
but can suffer from rapidly increasing variance 
as the horizon grows~\citep{precup2000eligibility,liu2020curse,jiang2016doublyrobustoffpolicyvalue}. 
Per-decision importance sampling alleviates this problem by applying 
only the likelihood ratios associated with the trajectory 
prefix relevant to each decision~\citep{precup2000eligibility}. 
Another widely used approach is to truncate or clip importance ratios.
Retrace truncates importance weights in multistep off-policy corrections to reduce variance~\citep{munos2016retrace}, while PPO clips the likelihood ratio in its surrogate objective to prevent excessively large policy updates~\citep{schulman2017ppo}.
Although their specific objectives differ, both methods limit the influence of extreme likelihood ratios on learning.
Ratio truncation therefore provides a practical trade-off: it introduces controlled bias in exchange for reduced variance and a more stable optimization signal.
ACER further combines truncated importance sampling with explicit bias correction and an efficient trust-region update to support stable off-policy actor--critic learning with experience replay~\citep{wang2017acer}.

\paragraph{Policy Optimization for Language Models.}
Reinforcement learning studies how an agent learns a policy that maximizes its expected cumulative reward through interaction with an environment~\citep{kaelbling1996reinforcement}. 
Policy optimization directly parameterizes this policy and improves it using gradients estimated from sampled trajectories~\citep{sutton1999policy}. 
Such methods have become a standard component of reinforcement-learning-based post-training for large language models (LLMs)~\citep{ouyang2022training,shao2024deepseekmathpushinglimitsmathematical}. 
Trust Region Policy Optimization constrains policy updates through  a KL-divergence trust region but requires an expensive constrained optimization procedure~\citep{schulman2015trpo}. 
PPO replaces this procedure with a clipped importance-ratio objective that supports multiple minibatch updates on the same rollout data~\citep{schulman2017ppo}. 
GRPO adapts this objective to LLM reinforcement learning by replacing the learned critic with group-relative advantages~\citep{shao2024deepseekmathpushinglimitsmathematical}, while DAPO introduces asymmetric clipping and other training techniques to improve exploration and large-scale optimization~\citep{yu2025dapo}. 
More recent methods reconsider the granularity and geometry of the clipping ratio. GSPO applies clipping to the geometric mean of the full-response likelihood ratio, improving the numerical stability of MoE reinforcement learning but assigning the same sequence-level statistic to every token~\citep{zheng2025gspo}. 
CTPO instead uses a cumulative prefix ratio and position-dependent log-space bounds, aligning the importance correction more closely with the causal structure of token-level gradients~\citep{zhang2026ctpo}. 
RIPO approaches clipping from a complementary geometric perspective and adapts the ratio bounds according to the behavior-policy probability, addressing the non-uniform policy divergence induced by fixed ratio intervals~\citep{cai2026ripo}. 
Our work combines prefix-level importance correction with a trust region constructed from the reverse-KL coordinate in its non-approximated nonlinear form, together with position-dependent budgets and behavior-probability calibration.

\section{Additional Derivations and Mathematical Properties}
\label{app:mathematical-properties}

\subsection{Conditions for the Prefix Reverse-KL Identity}
\label{app:reverse-kl-identity}

Fix a prompt \(x\) and a prefix length \(t\). The behavior policy \(\mu\) and the updated policy \(\pi\) induce normalized distributions over prefixes of length \(t\). We assume common support,
\[
\pi(o_{1:t}\mid x)>0
\quad\Longrightarrow\quad
\mu(o_{1:t}\mid x)>0,
\]
and finite reverse KL divergence. For variable-length responses, an absorbing end-of-sequence state can be used to define a normalized prefix distribution at every position; the distributions considered here are not conditioned on the response remaining unfinished.

Using the cumulative prefix ratio \( r_t^{\mathrm{pre}} = \pi(o_{1:t}\mid x) / \mu(o_{1:t}\mid x) \) its expectation under the behavior-prefix distribution is
\begin{equation}
\begin{aligned}
\mathbb E_{o_{1:t}\sim\mu(\cdot\mid x)}
\left[r_t^{\mathrm{pre}}\right]
&=
\sum_{o_{1:t}}
\mu(o_{1:t}\mid x)
\frac{\pi(o_{1:t}\mid x)}
     {\mu(o_{1:t}\mid x)}
\\
&=
\sum_{o_{1:t}}
\pi(o_{1:t}\mid x)
=1.
\end{aligned}
\label{eq:app-unit-expectation}
\end{equation}

The common-support assumption permits cancellation of the behavior probability throughout the support of the current policy. Starting from the definition of KL divergence, we can therefore express the prefix reverse KL as
\begin{equation}
\begin{aligned}
K_{x,t}(\mu,\pi)
&:=
D_{\mathrm{KL}}\!\left(
\mu(\cdot_{1:t}\mid x)
\,\middle\|\,
\pi(\cdot_{1:t}\mid x)
\right)
\\
&=
\sum_{o_{1:t}}
\mu(o_{1:t}\mid x)
\log
\frac{
\mu(o_{1:t}\mid x)
}{
\pi(o_{1:t}\mid x)
}
\\
&=
\mathbb E_{o_{1:t}\sim\mu(\cdot_{1:t}\mid x)}
\left[
\log
\frac{
\mu(o_{1:t}\mid x)
}{
\pi(o_{1:t}\mid x)
}
\right]
\\
&=
\mathbb E_{o_{1:t}\sim\mu(\cdot_{1:t}\mid x)}
\left[
-\log r_t^{\mathrm{pre}}
\right]
\\
&=
\mathbb E_{o_{1:t}\sim\mu(\cdot_{1:t}\mid x)}
\left[
-\log r_t^{\mathrm{pre}}
\right]
+
\mathbb E_{o_{1:t}\sim\mu(\cdot_{1:t}\mid x)}
\left[
r_t^{\mathrm{pre}}-1
\right]
\\
&=
\mathbb E_{o_{1:t}\sim\mu(\cdot_{1:t}\mid x)}
\left[
r_t^{\mathrm{pre}}
-1-\log r_t^{\mathrm{pre}}
\right]
\\
&=
\mathbb E_{o_{1:t}\sim\mu(\cdot_{1:t}\mid x)}
\left[
\phi\!\left(r_t^{\mathrm{pre}}\right)
\right],
\qquad
\phi(r)=r-1-\log r.
\end{aligned}
\label{eq:app-reverse-kl-identity}
\end{equation}
The resulting identity is an expectation-level statement: \(\phi(r_t^{\mathrm{pre}})\) is a nonnegative samplewise coordinate whose behavior-policy expectation equals the prefix reverse KL.

\subsection{Local Connection to Direct Ratio Clipping}
\label{app:local-ratio-clipping}

\paragraph{Local KL geometry and ratio bounds.}

For any positive likelihood ratio \(r\), expanding \(\log r\) around \(r=1\) gives
\begin{equation}
\begin{aligned}
\log r
&=
(r-1)-\frac{(r-1)^2}{2}
+O\!\left((r-1)^3\right),
\end{aligned}
\label{eq:app-ratio-expansion}
\end{equation}
Apply this expansion to the KL divergence:
\begin{equation}
\begin{aligned}
&D_{\mathrm{KL}}(\pi_{\mathrm{old}}(\cdot\mid s)\,\|\,\pi_\theta(\cdot\mid s)) \\
&=\mathbb E_{\pi_{\mathrm{old}}}\!\left[
-\log\frac{\pi_\theta(a_t\mid s_t)}{\pi_{\mathrm{old}}(a_t\mid s_t)}
\right]
&&\text{(definition of KL)}\\
&=\mathbb E_{\pi_{\mathrm{old}}}[-\log r_t]
&&\text{(definition of $r_t$)}\\
&=\mathbb E_{\pi_{\mathrm{old}}}[r_t-1-\log r_t]
&&\text{($\mathbb E_{\pi_{\mathrm{old}}}[r_t]=1$)}\\
&\approx\frac12\mathbb E_{\pi_{\mathrm{old}}}[(r_t-1)^2]
&&\text{(second-order expansion)},
\end{aligned}
\end{equation}
Replacing the nonlinear coordinate by its second-order approximation therefore converts a coordinate budget into an approximately symmetric ratio interval:
\begin{equation}
\frac{(r-1)^2}{2}\leq\delta
\quad\Longleftrightarrow\quad
1-\sqrt{2\delta}
\leq r\leq
1+\sqrt{2\delta}.
\label{eq:app-quadratic-ratio-constraint}
\end{equation}
For small \(\delta\), this recovers the form of the direct ratio interval used by PPO-style clipping, with \(\epsilon=\sqrt{2\delta}\). Separate lower- and upper-side budgets yield the approximate radii \(\sqrt{2\delta^-}\) and \(\sqrt{2\delta^+}\).

Applying this argument to \(r_t^{\mathrm{pre}}\) shows that direct prefix-ratio bounds correspond locally to a second-order approximation of the reverse-KL coordinate. \method{} instead determines its bounds from the nonlinear coordinate \(\phi(r_t^{\mathrm{pre}})\) without replacing it by the quadratic expression.

\paragraph{Relation between clipping and a hard KL constraint.}

A samplewise coordinate interval is stronger than an expectation-level KL budget, but optimizing a clipped objective does not project the resulting policy onto that interval. Clipping changes the optimization incentive once a sampled ratio crosses a boundary; it does not impose a hard constraint on every action probability.

For example, consider a two-action state with \(\mu=(1/2,1/2)\), advantages \((1,-1)\), and \(\pi=(1-\eta,\eta)\). For \(0<\epsilon<1\) and \(0<\eta\leq(1-\epsilon)/2\), both advantage-favored terms have reached their clipping plateaus, while
\begin{equation}
D_{\mathrm{KL}}(\mu\|\pi)
=
-\frac12\log\!\left(4\eta(1-\eta)\right)
\longrightarrow\infty
\quad\text{as}\quad
\eta\downarrow0.
\label{eq:app-ppo-kl-counterexample}
\end{equation}
Thus, clipping provides a tractable pessimistic objective rather than a hard KL projection~\citep{wang2020truly}.

\subsection{Existence and Properties of the Clipping Bounds}
\label{app:coordinate-properties}

\paragraph{Non-negativity and monotonicity.}

For \(r>0\),
\begin{equation}
\phi'(r)=1-\frac{1}{r},
\qquad
\phi''(r)=\frac{1}{r^2}>0.
\label{eq:app-coordinate-derivatives}
\end{equation}
Thus, \(\phi\) is strictly convex, decreasing on \((0,1)\), and increasing on \((1,\infty)\). Since \(\phi(1)=0\), it is nonnegative and has a unique minimum at one. Moreover,
\begin{equation}
\lim_{r\to0^+}\phi(r)=+\infty,
\qquad
\lim_{r\to+\infty}\phi(r)=+\infty.
\label{eq:app-coordinate-limits}
\end{equation}
Every positive finite coordinate budget therefore defines exactly one boundary on each side of one.

\paragraph{Calibrated cumulative-ratio bounds.}

With the calibrated budgets in Eq.~(\ref{eq:calibrated-prefix-budget}), the cumulative-ratio endpoints
are the unique solutions of
\begin{equation}
\phi(\ell_{i,t})
=
\frac{t\delta^-}{[\mu(o_{i,1:t}\mid x)]^{1/t}},
\quad
\phi(u_{i,t})
=
\frac{t\delta^+}{[\mu(o_{i,1:t}\mid x)]^{1/t}},
\qquad
0<\ell_{i,t}\leq1\leq u_{i,t}.
\label{eq:app-cumulative-bound-equations}
\end{equation}
The corresponding endpoints for the geometric-average ratio are \(\ell_{i,t}^{1/t}\) and \(u_{i,t}^{1/t}\). Because the \(t\)-th-root transformation is strictly increasing, it preserves both endpoint ordering and boundary-violation decisions.

\paragraph{Dependence on the budgets.}

Holding the sampled prefix and its behavior probability fixed, increasing \(\delta^-\) decreases \(\ell_{i,t}\), while increasing \(\delta^+\) increases \(u_{i,t}\). Increasing either budget therefore relaxes the corresponding side of the admissible interval. As either budget approaches zero, its cumulative endpoint and geometric-average endpoint approach one. As the lower-side budget tends to infinity, \(\ell_{i,t}\) approaches zero; as the upper-side budget tends to infinity, \(u_{i,t}\) grows without bound.

\section{Additional Implementation Details}
\label{app:surrogate-implementation}

\subsection{Geometric Transformation and Straight-Through Gradient}
\label{app:straight-through-gradient}

\paragraph{Transformation of the admissible interval.}
The geometric-average ratio is
\[
g_{i,t}
=
\left(r_{i,t}^{\mathrm{pre}}\right)^{1/t}.
\]
Since the \(t\)-th-root transformation is strictly increasing,
\begin{equation}
r_{i,t}^{\mathrm{pre}}\in[\ell_{i,t},u_{i,t}]
\quad\Longleftrightarrow\quad
g_{i,t}\in
\left[
\ell_{i,t}^{1/t},
u_{i,t}^{1/t}
\right].
\label{eq:app-boundary-equivalence}
\end{equation}
Thus, the transformation preserves which side of the admissible interval a prefix ratio violates, while expressing the optimization weight on a per-token scale.

\paragraph{Current-token-only gradient.}
Direct differentiation of \(g_{i,t}\) propagates through all current policy probabilities in the prefix:
\begin{equation}
\nabla_\theta g_{i,t}
=
\frac{g_{i,t}}{t}
\sum_{k=1}^{t}
\nabla_\theta
\log\pi_\theta(o_{i,k}\mid x,o_{i,<k}).
\label{eq:direct-geometric-prefix-gradient}
\end{equation}
To use prefix information in the forward weight while retaining a current-token-only gradient path, we construct
\begin{equation}
\widetilde g_{i,t}
=
\operatorname{sg}(g_{i,t})
\frac{
\pi_\theta(o_{i,t}\mid x,o_{i,<t})
}{
\operatorname{sg}\!\left(
\pi_\theta(o_{i,t}\mid x,o_{i,<t})
\right)
},
\label{eq:straight-through-prefix-ratio}
\end{equation}
where \(\operatorname{sg}\) denotes stop-gradient. Its forward value is
\(g_{i,t}\), whereas
\begin{equation}
\nabla_\theta\widetilde g_{i,t}
=
g_{i,t}
\nabla_\theta
\log\pi_\theta(o_{i,t}\mid x,o_{i,<t}).
\label{eq:straight-through-prefix-gradient}
\end{equation}
The cumulative prefix therefore determines the forward weight, while the differentiable path remains local to the current token. The advantages, behavior-policy probabilities, calibrated budgets, and transformed endpoints \(\ell_{i,t}^{1/t}\) and \(u_{i,t}^{1/t}\) are held fixed during differentiation.

\subsection{Sign-Aware Clipping Operator}
\label{app:sign-aware-surrogate}

For a positive weight \(q\), advantage \(A\), and interval
\(0<\ell\leq1\leq u\), we define the sign-aware clipping operator
\begin{equation}
\mathcal C(q,A;\ell,u)
=
A
\begin{cases}
\min(q,u), & A\geq0,\\[3pt]
\operatorname{clip}(q,\ell,u), & A<0.
\end{cases}
\label{eq:app-sign-aware-operator}
\end{equation}
For positive advantages, the operator removes the incentive to increase the weight beyond the upper endpoint. For negative advantages, it restricts the weight to the full admissible interval, including the additional upper-side cap.

The token-level contribution in the main objective is therefore
\begin{equation}
\mathcal S_{i,t}
=
\mathcal C\!\left(
\widetilde g_{i,t},
\widehat A_i;
\ell_{i,t}^{1/t},
u_{i,t}^{1/t}
\right).
\label{eq:sign-complete-surrogate}
\end{equation}
An equivalent nested representation is
\begin{equation}
\mathcal S_{i,t}
=
\begin{cases}
\displaystyle
\min\!\left\{
\widetilde g_{i,t}\widehat A_i,\,
\operatorname{clip}\!\left(
\widetilde g_{i,t},
\ell_{i,t}^{1/t},
u_{i,t}^{1/t}
\right)\widehat A_i
\right\},
& \widehat A_i\geq0,
\\[8pt]
\displaystyle
\max\!\left\{
\min\!\left\{
\widetilde g_{i,t}\widehat A_i,\,
\operatorname{clip}\!\left(
\widetilde g_{i,t},
\ell_{i,t}^{1/t},
u_{i,t}^{1/t}
\right)\widehat A_i
\right\},
u_{i,t}^{1/t}\widehat A_i
\right\},
& \widehat A_i<0.
\end{cases}
\label{eq:app-nested-surrogate}
\end{equation}

Away from the clipping boundaries, its gradient is
\begin{equation}
\begin{aligned}
\nabla_\theta\mathcal S_{i,t}
={}&
\widehat A_i g_{i,t}
\nabla_\theta
\log\pi_\theta(o_{i,t}\mid x,o_{i,<t})
\\
&\times
\begin{cases}
\mathbf 1\!\left\{
g_{i,t}<u_{i,t}^{1/t}
\right\},
& \widehat A_i\geq0,
\\[3pt]
\mathbf 1\!\left\{
\ell_{i,t}^{1/t}
<
g_{i,t}
<
u_{i,t}^{1/t}
\right\},
& \widehat A_i<0.
\end{cases}
\end{aligned}
\label{eq:app-sign-aware-gradient}
\end{equation}
The prefix determines the forward weight and the clipping decision, while the differentiable path remains local to the current token. The training loss is the negative of the objective in Eq.~(\ref{eq:exact-prefix-objective}).

\subsection{Reference Coding Implementation}
\label{app:reference-implementation}

Listing~\ref{lst:essencerl-loss} presents a compact implementation of the \method{} policy loss. The implementation computes cumulative prefix log-ratios, converts the calibrated cumulative-ratio bounds into the geometric-average endpoints \(\ell_{i,t}^{1/t}\) and \(u_{i,t}^{1/t}\), constructs the current-token-only straight-through weight, and applies the sign-aware operator in Eq.~(\ref{eq:app-sign-aware-operator}).

In the listing, after division by the prefix length, \texttt{lower} and \texttt{upper} denote $\log(\ell_{i,t}^{1/t})$ and $\log(u_{i,t}^{1/t})$, respectively. The inputs \texttt{logp}, \texttt{old\_logp}, \texttt{adv}, and \texttt{mask} have shape \([B,T]\). Diagnostic logging, alternative loss modes, and framework-specific distributed reductions are omitted.

\begin{lstlisting}[
    language=Python,
    basicstyle=\ttfamily\footnotesize,
    keywordstyle=\bfseries,
    commentstyle=\itshape,
    numbers=left,
    numberstyle=\tiny,
    numbersep=6pt,
    frame=single,
    breaklines=true,
    columns=fullflexible,
    keepspaces=true,
    showstringspaces=false,
    captionpos=t,
    caption={Implementation of the \method{} policy loss.},
    label={lst:essencerl-loss}
]
def solve_bounds(log_budget):
    """Solve exp(z) - 1 - z = budget on both sides of zero."""
    b = log_budget
    c = b.clamp(max=12.0).exp()
    q = (2.0 * c).sqrt()

    # Small-budget root expansions.
    lower_small = -(q + c / 3.0)
    upper_small = (q - c / 3.0).clamp_min(0.0)

    # Positive root: Newton iterations.
    upper = torch.where(c < 1.0, q, torch.log1p(c) + 0.5)
    for _ in range(4):
        d = torch.expm1(upper)
        upper = (
            upper - (d - upper - c) / d.clamp_min(1e-12)
        ).clamp_min(0.0)

    # Negative root: write z = -u, with u >= 0.
    u = torch.where(c < 1.0, q, c + 1.0)
    for _ in range(4):
        residual = torch.exp(-u) - 1.0 + u - c
        derivative = -torch.expm1(-u)
        u = (
            u - residual / derivative.clamp_min(1e-12)
        ).clamp_min(0.0)

    # Large-budget roots, evaluated using the log budget.
    upper_large = b.clamp_min(0.0)
    for _ in range(4):
        upper_large = torch.logaddexp(
            b, torch.log1p(upper_large)
        )
    lower_large = -(b.exp() + 1.0)
    lower = torch.where(c < 1e-4, lower_small, -u)
    upper = torch.where(c < 1e-4, upper_small, upper)
    lower = torch.where(b > 12.0, lower_large, lower)
    upper = torch.where(b > 12.0, upper_large, upper)
    return lower, upper


def epic_loss(
    logp, old_logp, adv, mask,
    delta_low=5e-4, delta_high=2e-3,
):
    # Prefix length and geometric-average log ratio.
    t = mask.cumsum(dim=-1).clamp_min(1.0)
    log_r = ((logp - old_logp) * mask).cumsum(dim=-1)
    log_g = log_r / t

    # Geometric-mean behavior probability calibrates the budget.
    log_p_old = (old_logp * mask).cumsum(dim=-1) / t
    log_budget_low = t.log() + math.log(delta_low) - log_p_old
    log_budget_high = t.log() + math.log(delta_high) - log_p_old

    # Convert cumulative-ratio bounds to geometric-average bounds.
    lower, _ = solve_bounds(log_budget_low)
    _, upper = solve_bounds(log_budget_high)
    lower, upper = lower / t, upper / t

    # Prefix-valued weight with a current-token-only gradient path.
    log_w = logp - logp.detach() + log_g.detach()
    log_w = log_w.clamp(-10.0, 10.0)
    log_w_clip = torch.minimum(
        torch.maximum(log_w, lower), upper
    ).clamp(-10.0, 10.0)

    # Pessimistic clipping and negative-advantage upper protection.
    loss = torch.maximum(
        -adv * log_w.exp(),
        -adv * log_w_clip.exp(),
    )
    negative_cap = -adv * upper.clamp(-10.0, 10.0).exp()
    loss = torch.where(
        adv < 0, torch.minimum(loss, negative_cap), loss
    )

    # Average tokens within each response, then average responses.
    lengths = mask.sum(dim=-1)
    response_loss = (loss * mask).sum(dim=-1) / (lengths + 1e-8)
    valid = (lengths > 0)
    return (response_loss * valid).sum() / valid.sum().clamp_min(1)
\end{lstlisting}

\FloatBarrier

\section{Experimental Configurations}
\label{app:experimental-config}

\subsection{Model and Training Framework}
\label{app:model-framework-resources}

All experiments use Qwen3-30B-A3B-Base ~\citep{yang2025qwen3technicalreport} \footnote{\url{https://huggingface.co/Qwen/Qwen3-30B-A3B-Base}}, a sparse mixture-of-experts language model.
We conduct reinforcement learning experiments using verl~\citep{sheng2024hybridflow}, with Megatron~\citep{shoeybi2020megatronlmtrainingmultibillionparameter} as the policy-training backend and vLLM~\citep{kwon2023efficientmemorymanagementlarge} as the rollout-generation backend.
Experiments are conducted using 32 NVIDIA A100 GPUs.

The main comparison evaluates GRPO, DAPO, GSPO, CTPO, RIPO, and \method{} under the same training and evaluation protocol ~\citep{shao2024deepseekmathpushinglimitsmathematical,
yu2025dapo,zheng2025gspo,zhang2026ctpo,cai2026ripo}.
The methods share the base model, training data, sampling settings, and optimization schedule, while their clipping parameters follow their respective definitions.
We additionally evaluate routing-replay variants under the configuration described in Section~\ref{app:routing-replay-config}.

\subsection{Datasets and Reward Verification}
\label{app:datasets-verification}

\paragraph{Training data and verification.}
We use a processed version of DAPO-Math-17K%
\footnote{\url{https://huggingface.co/datasets/BytedTsinghua-SIA/DAPO-Math-17k}},
containing 17,917 training entries~\citep{yu2025dapo}.
A rule-based verifier compares the normalized final answer with
the reference answer, assigning a correctness score of 1 for a
correct response and 0 otherwise.
Evaluation uses the same binary correctness scores.

\paragraph{Evaluation benchmarks.}
The evaluation suite contains 130 distinct prompts:
30 from AIME 2024%
\footnote{\url{https://huggingface.co/datasets/HuggingFaceH4/aime_2024}},
30 from AIME 2025%
\footnote{\url{https://huggingface.co/datasets/math-ai/aime25}},
40 from AMC 2023%
\footnote{\url{https://huggingface.co/datasets/math-ai/amc23}},
and 30 from HMMT February 2025%
\footnote{\url{https://huggingface.co/datasets/MathArena/hmmt_feb_2025}}.
For AMC 2023, we remove repeated entries with identical prompt
text so that each retained problem is evaluated once.

\begin{table}[!htbp]
    \centering
    \caption{
        Configurations of the main Qwen3-30B-A3B-Base experiments.
        Batch and minibatch sizes are measured in prompts.
    }
    \label{tab:shared-training-config}
    \small
    \setlength{\tabcolsep}{3pt}
    \renewcommand{\arraystretch}{1.08}

    \begin{tabularx}{\linewidth}{
        @{}l*{6}{>{\centering\arraybackslash}X}@{}
    }
        \toprule
        Parameter
        & GRPO & DAPO & GSPO & CTPO & RIPO
        & \method{} \\
        \midrule

        \multicolumn{7}{l}{
            \textit{Sampling and sequence limits}
        } \\
        Temperature (train / val)
        & 1.0 / 1.0 & 1.0 / 1.0 & 1.0 / 1.0
        & 1.0 / 1.0 & 1.0 / 1.0 & 1.0 / 1.0 \\
        Top-$p$ (train / val)
        & 1.0 / 0.7 & 1.0 / 0.7 & 1.0 / 0.7
        & 1.0 / 0.7 & 1.0 / 0.7 & 1.0 / 0.7 \\
        Responses per prompt (train / val)
        & 8 / 8 & 8 / 8 & 8 / 8
        & 8 / 8 & 8 / 8 & 8 / 8 \\
        Oversampling ratio
        & 0.1 & 0.1 & 0.1 & 0.1 & 0.1 & 0.1 \\
        Maximum prompt length (tokens)
        & 2,048 & 2,048 & 2,048
        & 2,048 & 2,048 & 2,048 \\
        Maximum response length (tokens)
        & 16,384 & 16,384 & 16,384
        & 16,384 & 16,384 & 16,384 \\
        \midrule

        \multicolumn{7}{l}{\textit{Clipping}} \\
        Clipping parameter (low)
        & $0.2$ & $0.2$ & $3{\times}10^{-4}$
        & $0.025$ & $0.05$ & $5{\times}10^{-4}$ \\
        Clipping parameter (high)
        & $0.2$ & $0.28$ & $4{\times}10^{-4}$
        & $0.05$ & $0.05$ & $2{\times}10^{-3}$ \\
        \midrule

        \multicolumn{7}{l}{
            \textit{Batching and optimization}
        } \\
        Rollout batch / minibatch size
        & 64 / 8 & 64 / 8 & 64 / 8
        & 64 / 8 & 64 / 8 & 64 / 8 \\
        PPO epochs / update depth
        & 1 / 8 & 1 / 8 & 1 / 8
        & 1 / 8 & 1 / 8 & 1 / 8 \\
        Actor learning rate
        & $10^{-6}$ & $10^{-6}$ & $10^{-6}$
        & $10^{-6}$ & $10^{-6}$ & $10^{-6}$ \\
        Learning-rate warmup steps
        & 10 & 10 & 10 & 10 & 10 & 10 \\
        Gradient-norm clipping threshold
        & 1.0 & 1.0 & 1.0 & 1.0 & 1.0 & 1.0 \\
        KL-loss coefficient
        & $10^{-3}$ & $10^{-3}$ & $10^{-3}$
        & $10^{-3}$ & $10^{-3}$ & $10^{-3}$ \\
        Entropy-bonus coefficient
        & 0 & 0 & 0 & 0 & 0 & 0 \\
        Total global training steps
        & 300 & 300 & 300 & 300 & 300 & 300 \\
        \bottomrule
    \end{tabularx}

    \vspace{3pt}
    \begin{minipage}{\linewidth}
        \footnotesize
        Clipping parameters follow their method-specific definitions
        and are not numerically interchangeable.
        For \method{}, the lower- and upper-side parameters are
        the coordinate budgets $\delta^-$ and $\delta^+$.
    \end{minipage}
\end{table}

\FloatBarrier

\subsection{Shared Training Configuration}
\label{app:shared-training-configuration}

Table~\ref{tab:shared-training-config} summarizes the sampling
settings, sequence limits, method-specific clipping parameters,
and optimization schedule.
Batch and minibatch sizes are measured in prompts, with eight
responses generated for each prompt.
A rollout batch of 64 prompts therefore contains 512 responses
and is divided into eight minibatches of 64 responses each.
Each rollout batch is used for one PPO epoch, corresponding to
an update depth of eight.

All main runs are trained for 300 global steps with an actor
learning rate of $10^{-6}$ and 10 learning-rate warmup steps.
The maximum prompt and response lengths are 2,048 and 16,384
tokens, respectively.
The gradient-norm clipping threshold is 1.0, the KL-loss
coefficient is $10^{-3}$, and no entropy bonus is applied.

\subsection{Evaluation and Checkpoint Selection}
\label{app:evaluation-protocol}

\paragraph{Evaluation schedule and metric.}
We evaluate the initial model before training and repeat evaluation every five global steps.
Each evaluation samples eight responses per prompt, producing 1,040 responses across the four benchmarks.
The evaluation temperature is 1.0 and top-$p$ is 0.7.
For each benchmark, $\mathrm{mean@8}$ is the average binary correctness over its prompts and their eight sampled responses.

\paragraph{Checkpoint selection and seed aggregation.}
We first compute $\mathrm{mean@8}$ within each benchmark and then take the equal-weight arithmetic mean of the four benchmark scores.
The four benchmarks receive equal weight regardless of their numbers of prompts.
For each run, we select the checkpoint that maximizes this four-benchmark average.
All benchmark scores reported for that run therefore come from the same selected checkpoint, rather than independently selected benchmark-specific checkpoints.
The main six-method comparison is repeated with three random seeds.
We apply checkpoint selection separately to each run and then average the selected-checkpoint scores across the three seeds.

\subsection{Routing-Replay Comparisons}
\label{app:routing-replay-config}

We evaluate R2 and R3 with a GRPO Clip-Higher objective using
$\epsilon_{\mathrm{low}}=0.2$ and
$\epsilon_{\mathrm{high}}=0.28$.
Both variants follow the shared configuration in
Table~\ref{tab:shared-training-config}.

R2 records expert assignments during behavior-policy
recomputation in the training engine and reuses them in
subsequent policy updates.
R3 records expert assignments during rollout generation and
replays them during behavior-policy recomputation and policy
updates%
~\citep{zheng2025gspo,zheng2025stabilizingrl,ma2025r3}.
\method{} does not use routing replay.

All routing-replay comparisons use the same four-benchmark
evaluation and per-run checkpoint-selection protocol described
in Section~\ref{app:evaluation-protocol}.
\FloatBarrier
\suppressfloats[t]

\section{Further Analysis}
\label{app:further-analysis}

\subsection{Validation Trajectories on Individual Benchmarks}
\label{app:validation-trajectories}

\begin{figure*}[!htbp]
    \centering
    \captionsetup[subfigure]{position=top,skip=2pt}

    \includegraphics[width=0.70\textwidth]
    {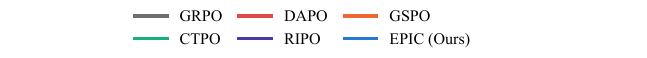}

    \vspace{3pt}

    \begin{subfigure}[t]{0.245\textwidth}
        \centering
        \caption{AIME 2024.}
        \label{fig:validation:aime24}
        \includegraphics[width=\linewidth]
        {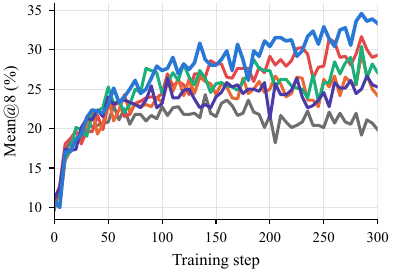}
    \end{subfigure}
    \hfill
    \begin{subfigure}[t]{0.245\textwidth}
        \centering
        \caption{AIME 2025.}
        \label{fig:validation:aime25}
        \includegraphics[width=\linewidth]
        {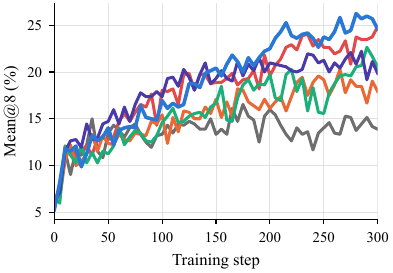}
    \end{subfigure}
    \hfill
    \begin{subfigure}[t]{0.245\textwidth}
        \centering
        \caption{AMC 2023.}
        \label{fig:validation:amc23}
        \includegraphics[width=\linewidth]
        {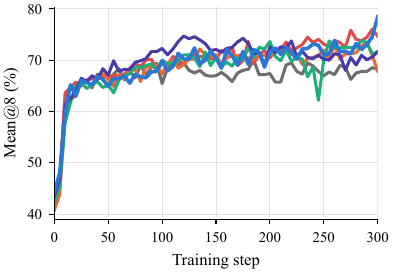}
    \end{subfigure}
    \hfill
    \begin{subfigure}[t]{0.245\textwidth}
        \centering
        \caption{HMMT 2025.}
        \label{fig:validation:hmmt25}
        \includegraphics[width=\linewidth]
        {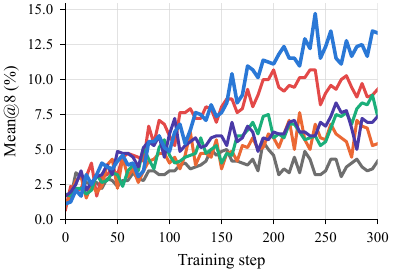}
    \end{subfigure}

    \caption{
        Validation \(\mathrm{mean@8}\) on the four individual benchmarks throughout training.
    }
    \label{fig:main-validation-curves}
\end{figure*}

Figure~\ref{fig:main-validation-curves} presents the complete validation
trajectories on individual benchmarks. Figure~\ref{fig:training-dynamics}\subref{fig:dynamics:validation}
reports their equal-weight average. \method{} develops a clear advantage in the
four-benchmark average during training and reaches a substantially
higher peak than the compared objectives. DAPO is the strongest
baseline and continues improving late in training, while CTPO and GSPO
reach similar aggregate peaks and RIPO remains below them. This
comparison suggests that the benefit of \method{} does not arise
from the choice of a prefix ratio alone; constructing and calibrating
the corresponding prefix-level trust region is also important.

\FloatBarrier

\subsection{Analysis of Training Diagnostics}
\label{app:training-diagnostics-analysis}

\begin{figure*}[!htbp]
    \centering
    \captionsetup[subfigure]{
        position=top,
        skip=2pt,
        font=small
    }

    \includegraphics[width=0.70\textwidth]
    {figures/off8_seed42_training_dynamics_panels/legend.pdf}

    \vspace{3pt}

    \begin{subfigure}[t]{0.245\textwidth}
        \centering
        \caption{Clip fraction.}
        \label{fig:diagnostics:clip}
        \includegraphics[width=\linewidth]
        {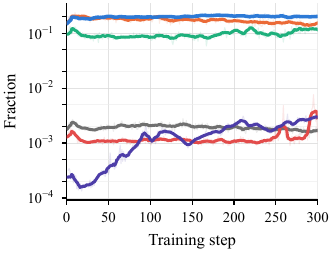}
    \end{subfigure}
    \hfill
    \begin{subfigure}[t]{0.245\textwidth}
        \centering
        \caption{Policy entropy.}
        \label{fig:diagnostics:entropy}
        \includegraphics[width=\linewidth]
        {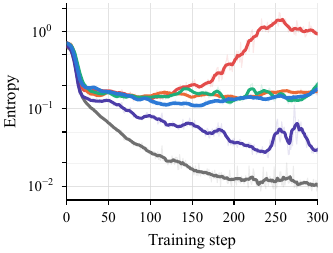}
    \end{subfigure}
    \hfill
    \begin{subfigure}[t]{0.245\textwidth}
        \centering
        \caption{Gradient norm.}
        \label{fig:diagnostics:grad}
        \includegraphics[width=\linewidth]
        {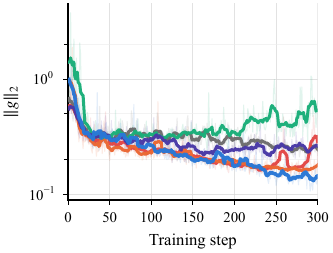}
    \end{subfigure}
    \hfill
    \begin{subfigure}[t]{0.245\textwidth}
        \centering
        \caption{Train--rollout gap.}
        \label{fig:diagnostics:gap}
        \includegraphics[width=\linewidth]
        {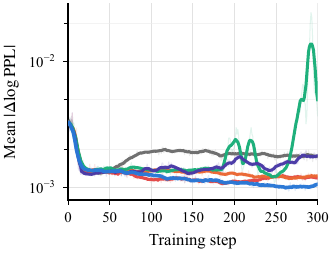}
    \end{subfigure}

    \caption{
        Additional training dynamics under the aligned off-policy MoE setting. 
        Curves average three seeds at matching training steps.
        Clip fraction, policy entropy, gradient norm, and train--rollout
        gap are shown on logarithmic vertical scales.
    }
    \label{fig:training-diagnostics}
\end{figure*}

Figure~\ref{fig:training-diagnostics} provides additional diagnostics for the optimization dynamics. The clip-fraction curves in Figure~\ref{fig:training-diagnostics}\subref{fig:diagnostics:clip} show that \method{} maintains a persistently active clipping regime throughout the later stages of training, with a clip fraction on the same order as GSPO. In contrast, GRPO, DAPO, and RIPO operate with substantially smaller late-stage clip fractions. This indicates that the prefix-level trust region continues to influence the update rather than becoming effectively inactive.

The entropy trajectories in Figure~\ref{fig:training-diagnostics}\subref{fig:diagnostics:entropy} show different degrees of policy concentration across methods.
GRPO and RIPO gradually reach entropy values on the order of $10^{-2}$, indicating a strong reduction in output diversity.
DAPO instead enters a higher and more variable entropy regime during the later stages of training. \method{} remains in an intermediate entropy regime, suggesting that its improved performance is not obtained by either excessive entropy collapse or uncontrolled entropy growth.

The gradient-norm curves in Figure~\ref{fig:training-diagnostics}\subref{fig:diagnostics:grad} indicate that all methods experience a rapid reduction in update magnitude during the initial stage. 
After this transient, \method{} maintains a comparatively controlled gradient scale and does not exhibit the sustained late-stage growth observed for some baselines.
Thus, its active clipping regime is not accompanied by excessively large parameter updates.

\FloatBarrier

Finally, Figure~\ref{fig:training-diagnostics}\subref{fig:diagnostics:gap}
shows the mean absolute train--rollout log-probability gap.
After the initial transient, the gap remains at a small numerical scale
for most methods, while \method{} maintains one of the lowest and most
stable late-stage values. This behavior is consistent with the intended
role of the prefix-aware objective: it regulates policy changes under
rollout reuse without requiring the additional routing-alignment
mechanisms used by replay-based approaches. Taken together, the four
diagnostics show that \method{} combines persistent trust-region
activity, controlled update magnitudes, non-collapsed entropy, and close
train--rollout consistency.

\subsection{Behavior Outside the Local-Approximation Regime}
\label{app:local-approximation-regime}

We examine whether the cumulative prefix ratios observed in our main
\method{} experiment remain close enough to one to justify a local
quadratic approximation. Let \(s=\log r^{\mathrm{pre}}_{i,t}\), and let \(m_j\) denote the mean of \(|s|\) over valid response tokens at training step \(j\).
The distribution of these step-level aggregates must be distinguished
from the distribution of individual token-level log-ratios.

\begin{figure}[!htbp]
    \centering
    \captionsetup[subfigure]{position=top,skip=2pt}
    \begin{subfigure}[t]{0.48\textwidth}
        \centering
        \caption{Observed prefix log-ratio magnitude.}
        \label{fig:local-regime:magnitude}
        \includegraphics[width=\linewidth]
        {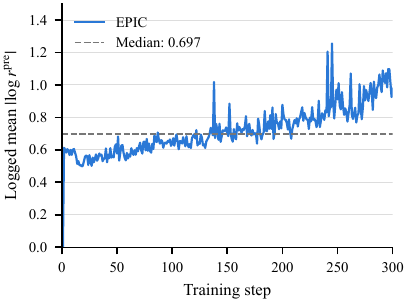}
    \end{subfigure}\hfill
    \begin{subfigure}[t]{0.48\textwidth}
        \centering
        \caption{Error at representative magnitudes.}
        \label{fig:local-regime:error}
        \includegraphics[width=\linewidth]
        {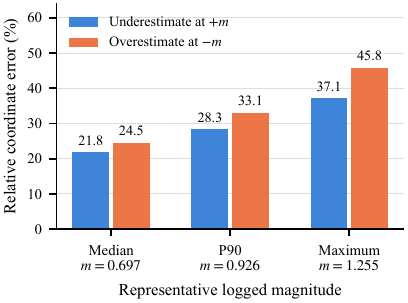}
    \end{subfigure}
    \caption{Local-approximation diagnostics for the main \method{}
    run. \textbf{(a)} Logged mean absolute cumulative prefix log-ratio at each
    training step; the dashed line marks its median across steps.
    \textbf{(b)} Relative error of the quadratic coordinate \(s^2/2\), evaluated
    at positive and negative representative magnitudes from panel (a),
    with \(\exp(s)-1-s\) as the reference denominator.
    These are analytical evaluations at representative magnitudes,
    not measured averages of token-level approximation errors.}
    \label{fig:local-regime}
\end{figure}

\paragraph{Observed cumulative deviations.}
Figure~\ref{fig:local-regime}\subref{fig:local-regime:magnitude}
shows that the logged magnitude does not remain near zero.
Across training steps, its median is \(0.697\), its 90th percentile
is \(0.926\), and its maximum is \(1.255\), reached at step 245.
The mean over steps 201--300 is \(0.873\), and 298 of the 300
recorded step-level aggregates exceed \(0.5\).
These observations indicate that cumulative prefix deviations are not
uniformly confined to an infinitesimal neighborhood of
\(s=0\), motivating a trust-region coordinate that does not require
this local assumption. 

\paragraph{Implications for the divergence coordinate.}
Figure~\ref{fig:local-regime}\subref{fig:local-regime:error}
evaluates the discrepancy between \(\exp(s)-1-s\) and \(s^2/2\)
at \(s=\pm m\), using the median, 90th-percentile, and maximum
logged magnitudes as reference points.
At the median magnitude, the quadratic approximation underestimates
the positive-side coordinate by \(21.8\%\) and overestimates the
negative-side coordinate by \(24.5\%\). These discrepancies increase
to \(28.3\%\) and \(33.1\%\) at the 90th-percentile magnitude,
and to \(37.1\%\) and \(45.8\%\) at the maximum magnitude.
For a fixed coordinate budget, the quadratic approximation therefore
permits a looser upper ratio bound and imposes a tighter lower ratio
bound than the unapproximated coordinate.
Retaining the full coordinate avoids this approximation-induced
distortion of the two sides of the trust region.
The observed magnitudes and the associated analytical discrepancies
support this design motivation, but do not by themselves isolate its
contribution to benchmark performance. In particular, evaluating a
nonlinear coordinate at a logged aggregate does not recover the
average coordinate or approximation error over individual tokens.

\FloatBarrier

\subsection{Ablation of Prefix-Probability Calibration}
\label{app:geomprob-ablation}

We disentangle two design choices in \method{}:
the behavior-policy probability used to calibrate the trust-region
budget and the scale used to represent the prefix ratio in the
optimization objective. The three configurations are summarized in
Table~\ref{tab:geomprob-ablation-config}.

The Raw-prob ablation replaces the geometric-mean behavior probability
in the budget calibration with the unnormalized cumulative probability
$\mu(o_{i,1:t}\mid x)$.
The Cumulative-ratio ablation keeps the geometric-mean probability
calibration but uses the cumulative prefix ratio directly in the
optimization objective.
The final row is the default \method{} configuration, which applies
geometric averaging to both the behavior-policy probability and the
prefix optimization ratio.

\begin{figure*}[!htbp]
    \centering
    \captionsetup[subfigure]{position=top,skip=2pt}

    \begin{subfigure}[t]{0.32\textwidth}
        \centering
        \caption{AIME24 validation.}
        \includegraphics[width=\linewidth]
        {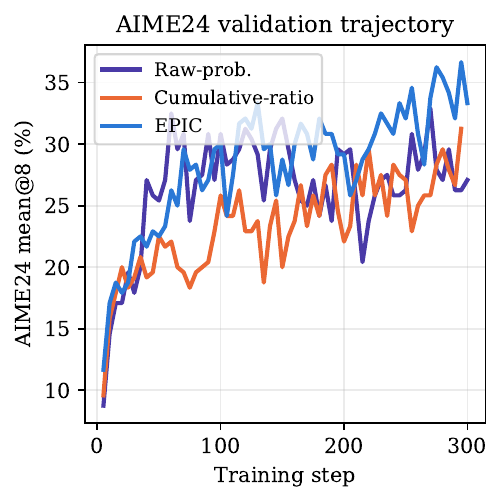}
    \end{subfigure}
    \hfill
    \begin{subfigure}[t]{0.32\textwidth}
        \centering
        \caption{Gradient norm.}
        \includegraphics[width=\linewidth]
        {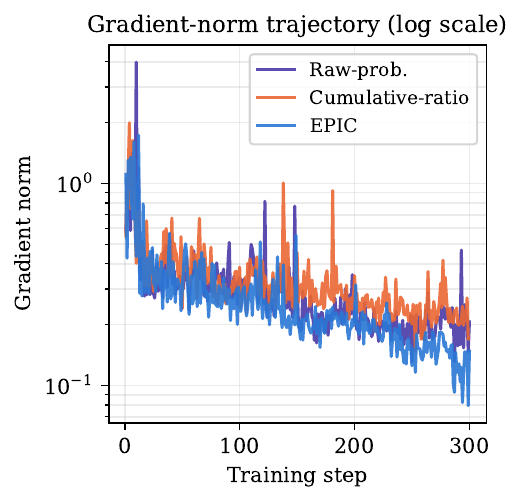}
    \end{subfigure}
    \hfill
    \begin{subfigure}[t]{0.32\textwidth}
        \centering
        \caption{Position-wise clipping.}
        \includegraphics[width=\linewidth]
        {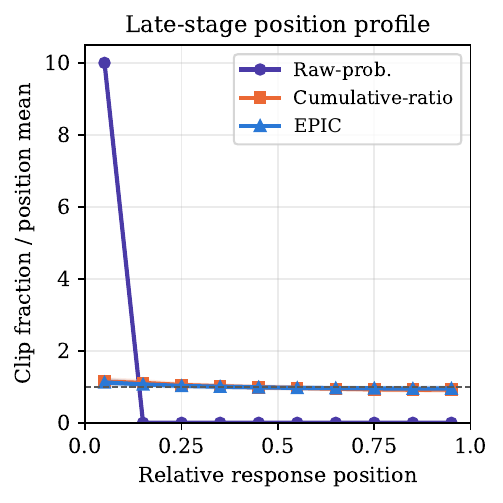}
    \end{subfigure}

    \caption{
        Ablation of prefix-probability calibration. The three panels
        show AIME24 validation performance, gradient-norm trajectories
        on a logarithmic scale, and late-stage position-wise clipping
        profiles. In the right panel, clipping fractions are normalized
        by the mean across position bins at each training step; shaded
        regions indicate the standard deviation over steps 201--300.
    }
    \label{fig:geomprob-ablation}
\end{figure*}

The validation trajectories in Figure~\ref{fig:geomprob-ablation}
show a clear difference between raw and geometric probability calibration.
Raw-prob reaches approximately $27.1\%$ on the final AIME24 evaluation,
compared with approximately $31.3\%$ for Cumulative-ratio and
$33.3\%$ for \method{}.

The gradient trajectories provide evidence that raw prefix
probabilities can induce less stable optimization.
In particular, Raw-prob exhibits a larger early gradient spike,
reaching a maximum gradient norm of approximately $3.98$,
compared with approximately $2.00$ for Cumulative-ratio and
approximately $1.73$ for \method{}.
This behavior is consistent with the strong length dependence
of the unnormalized cumulative prefix probability.

The position-wise profiles show the mechanism behind this difference.
With Raw-prob, clipping is concentrated almost entirely in the first
relative-position bin and becomes negligible afterward.
In contrast, Cumulative-ratio and \method{} remain close to their
position-wise mean across the response.
Thus, replacing the raw cumulative probability with a geometric
average removes the severe position imbalance induced by the
exponential length dependence of the prefix probability.

\begin{table}[!htbp]
    \centering
    \small
    \setlength{\tabcolsep}{5pt}
    \renewcommand{\arraystretch}{1.08}
    \caption{
        Configurations used in the prefix-probability ablation.
    }
    \label{tab:geomprob-ablation-config}
    \begin{tabular}{lll}
        \toprule
        Configuration & Budget calibration & Optimization ratio \\
        \midrule
        Raw-prob
        &
        $\displaystyle
        \frac{t\delta^\pm}{\mu(o_{i,1:t}\mid x)}$
        &
        $\displaystyle
        \left[r^{\mathrm{pre}}_{i,t}(\theta)\right]^{1/t}
        $
        \\[8pt]

        Cumulative-ratio
        &
        $\displaystyle
        \frac{t\delta^\pm}
        {\left[\mu(o_{i,1:t}\mid x)\right]^{1/t}}$
        &
        $\displaystyle
        r^{\mathrm{pre}}_{i,t}(\theta)
        $
        \\[8pt]

        \method{} (default)
        &
        $\displaystyle
        \frac{t\delta^\pm}
        {\left[\mu(o_{i,1:t}\mid x)\right]^{1/t}}$
        &
        $\displaystyle
        \left[r^{\mathrm{pre}}_{i,t}(\theta)\right]^{1/t}
        $
        \\
        \bottomrule
    \end{tabular}
\end{table}
\FloatBarrier

\FloatBarrier

\subsection{Memory Cost of Routing-Replay State}
\label{app:routing-replay-memory}

Routing replay introduces additional state beyond the token IDs,
log-probabilities, and rewards required by the policy objective.
For each retained token, it records the selected experts at every
MoE layer so that the routing decisions can be reused during
subsequent policy evaluations.
In the historical implementation associated with our experiments,
R2 converts the recorded expert indices to \texttt{uint8} before
transferring them to CPU memory.
The vLLM backport used for R3 instead stores the rollout routing
indices as \texttt{int32}, including its host shared-memory buffer.
The subsequent agent-loop processing preserves this dtype when
constructing padded routing tensors.

\paragraph{Storage estimate.}
For a batch containing \(N\) responses, each stored at length \(T\),
the memory required by one dense tensor of routing indices is
\begin{equation}
    M_{\mathrm{indices}}
    = N T L_{\mathrm{MoE}} K b,
    \label{eq:router-index-memory}
\end{equation}
where \(L_{\mathrm{MoE}}\) is the number of MoE layers, \(K\) is the number of selected experts per token, and \(b\) is the number of bytes per index. Qwen3-30B-A3B has 48 MoE layers and selects eight experts per token. Our rollout batch contains \(64\times8=512\) responses, and the configured maximum prompt-plus-response length is \(2{,}048+16{,}384=18{,}432\) tokens. Table~\ref{tab:router-index-memory} reports the corresponding single-copy storage estimates. For the same tensor shape, the R3 rollout representation requires four times as much storage as the compressed R2 representation.

\paragraph{Relation to observed process memory.}
Beyond the single-tensor storage estimated in Table~\ref{tab:router-index-memory}, the process-memory differences in the main-text figure also reflect shared-memory mappings, temporary storage during routing-index collection and concatenation, and larger routing tensors produced during validation.
The R3 rollout backend copies routing indices out of shared memory; the agent loop then constructs padded tensors and concatenates them within workers, followed by batch aggregation in the training coordinator.
R2 likewise performs gathering, CPU transfer, and concatenation of recorded indices. 
Inputs and newly allocated outputs may coexist during these operations, increasing transient memory usage.
R3 also allocates a host shared-memory buffer according to the configured KV-token capacity, whose size need not equal that of the current response batch.
These storage and data-movement mechanisms provide a concrete explanation for why routing replay can incur substantial CPU-memory overhead, consistent with the process-memory differences in Figure~\ref{fig:routing-replay:memory}.
\begin{table}[!htbp]
    \centering
    \caption{
        Estimated storage for one dense routing-index tensor.
    }
    \label{tab:router-index-memory}
    \small
    \setlength{\tabcolsep}{6pt}
    \renewcommand{\arraystretch}{1.10}
    \begin{tabular}{lccrr}
        \toprule
        & & & \multicolumn{2}{c}{Single-copy storage (GiB)} \\
        \cmidrule(lr){4-5}
        Routing state & Index dtype & Bytes/index
        & \(T=9{,}000\) & \(T=18{,}432\) \\
        \midrule
        R2 CPU indices
        & \texttt{uint8} & 1 & 1.65 & 3.375 \\
        R3 rollout indices
        & \texttt{int32} & 4 & 6.59 & 13.50 \\
        \bottomrule
    \end{tabular}
\end{table}
\FloatBarrier

\end{document}